\documentclass[journal]{IEEEtran}
\usepackage{cite}
\usepackage{amsmath}
\usepackage{amsfonts}
\usepackage{amssymb}
\usepackage{algorithm}
\usepackage{algorithmic}
\usepackage{array}
\usepackage{bm}
\usepackage{graphicx}
\usepackage{cleveref}
\usepackage{footmisc}
\usepackage{color}
\usepackage[labelformat=simple]{subcaption}
\usepackage{multirow}
\usepackage{booktabs}
\usepackage{enumerate}
\usepackage{graphbox} 

\definecolor{orange}{rgb}{1,0.3,0}

\newcolumntype{C}[1]{>{\centering\arraybackslash}p{#1}}

\newcommand{\eg}{\textit{e.g. }}
\newcommand{\ie}{\textit{i.e. }}
\newcommand{\etal}{\textit{et al. }}
\renewcommand{\subsubsection}[1]{{\textbf{#1. }}}

\begin{document}

\title{3D Quasi-Recurrent Neural Network for Hyperspectral Image Denoising}

\author{Kaixuan Wei, Ying Fu,~\IEEEmembership{Member,~IEEE}, and Hua
  Huang,~\IEEEmembership{Senior Member,~IEEE}}

\markboth{IEEE Transactions on Neural Networks and Learning System}%
{Shell \MakeLowercase{\textit{et al.}}: Bare Demo of IEEEtran.cls for IEEE Journals}
\maketitle
\begin{abstract}
  In this paper, we propose an alternating directional 3D quasi-recurrent neural
  network for hyperspectral image (HSI) denoising, which can effectively embed
  the domain knowledge --- structural spatio-spectral correlation and global
  correlation along spectrum. Specifically, 3D convolution is utilized to
  extract structural spatio-spectral correlation in an HSI, while a
  quasi-recurrent pooling function is employed to capture the global correlation
  along spectrum. Moreover, alternating directional
  structure 
  is introduced to eliminate the causal dependency with no additional
  computation cost. The proposed model is capable of modeling spatio-spectral
  dependency while preserving the flexibility towards HSIs with arbitrary number
  of bands. 
  Extensive experiments on HSI denoising demonstrate
  significant improvement over state-of-the-arts under various noise settings,
  in terms of both restoration accuracy and computation time. Our code is available at https://github.com/Vandermode/QRNN3D. 
\end{abstract}
\IEEEpeerreviewmaketitle

\section{Introduction}
\IEEEPARstart{H}{yperspectral} image (HSI) is made up of massive discrete wavebands for each spatial position of real scenes and provides much richer information about
scenes than RGB images, which has led to numerous applications in
remote sensing \cite{lillesand2014remote,thenkabail2016hyperspectral},
classification \cite{camps2014advances,Wang2017Salient,Ping2014Jointly,Akhtar2018Nonparametric,Yang2018Self}, tracking \cite{van2010tracking}, face
recognition \cite{uzair2015HSIface}, and more. However, due to the limited light
for each band, traditional HSIs are often degraded by various noises (i.e.,
Gaussian, stripe, deadline, and impulse noises) during the acquisition process.
These degradations negatively influence the performance of all subsequent HSI
processing tasks aforementioned. Therefore, HSI denoising is an essential
pre-processing in the typical workflow of HSI analysis and processing.

Recently, more HSI denoising works pay attention to the domain knowledge of the
HSI --- structural spatio-spectral correlation and global correlation along
spectrum (GCS) \cite{xie2016multispectral}. Top-performing classical methods
\cite{chang2017hyper,xie2016multispectral,chang2017weighted,wang2017hyperspectral,WEI2019412}
typically utilize non-local low-rank tensors to model them. Although these
methods achieve higher accuracy by effectively considering these underlying
characteristics, the performance of such methods is inherently determined by how
well the human handcrafted prior (\eg low-rank tensors) matches with the
intrinsic characteristics of an HSI. Besides, such approaches generally
formulate the HSI denoising as a complex optimization problem to be solved iteratively, making the denoising process time-consuming.

\begin{figure}[!htbp]
\centering
\begin{subfigure}[b]{.45\linewidth}
\centering
\includegraphics[width=1\linewidth,clip,keepaspectratio]{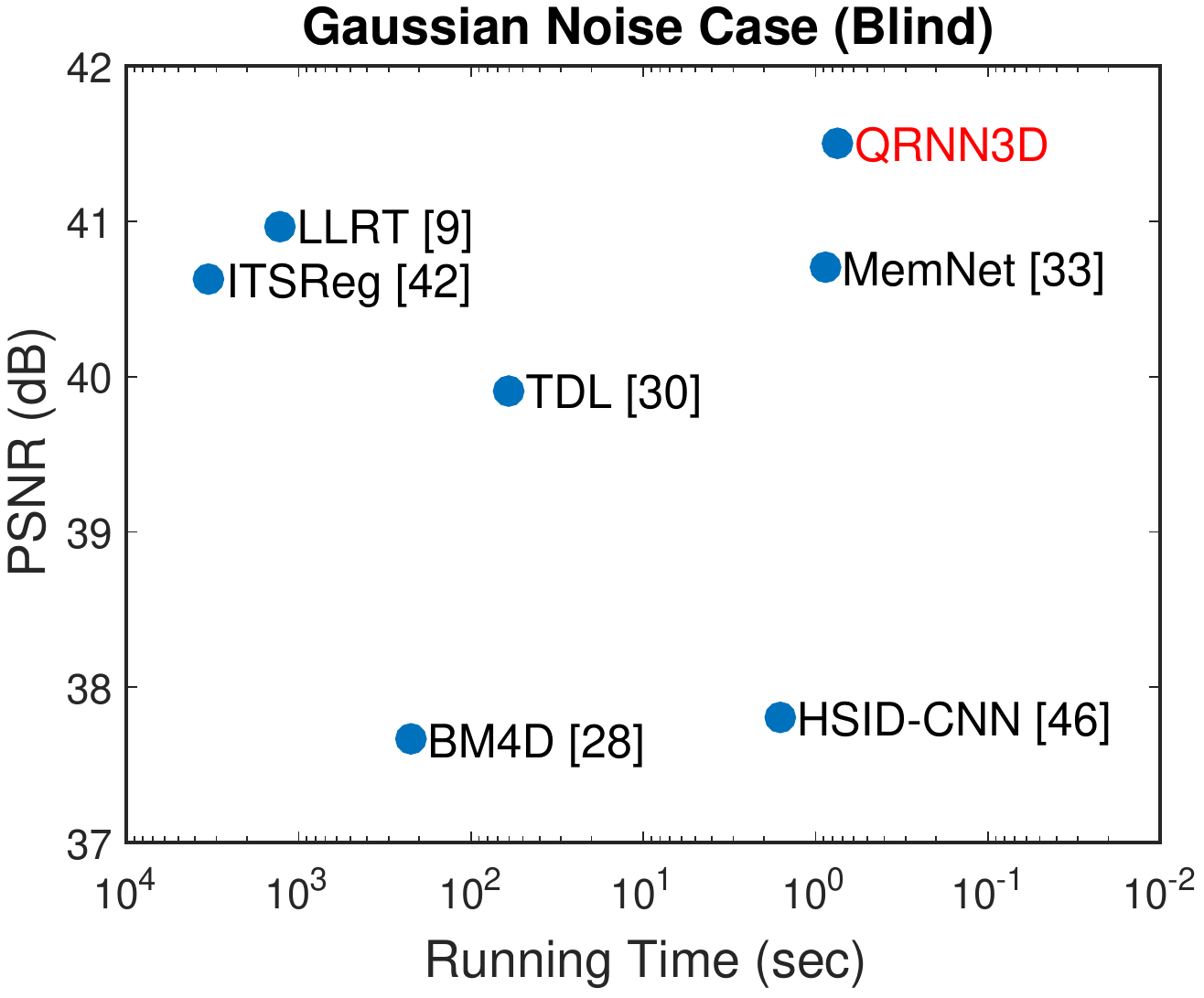}
\label{fig:runtime:1}
\end{subfigure}
\begin{subfigure}[b]{.45\linewidth}
  \centering
\includegraphics[width=1\linewidth,clip,keepaspectratio]{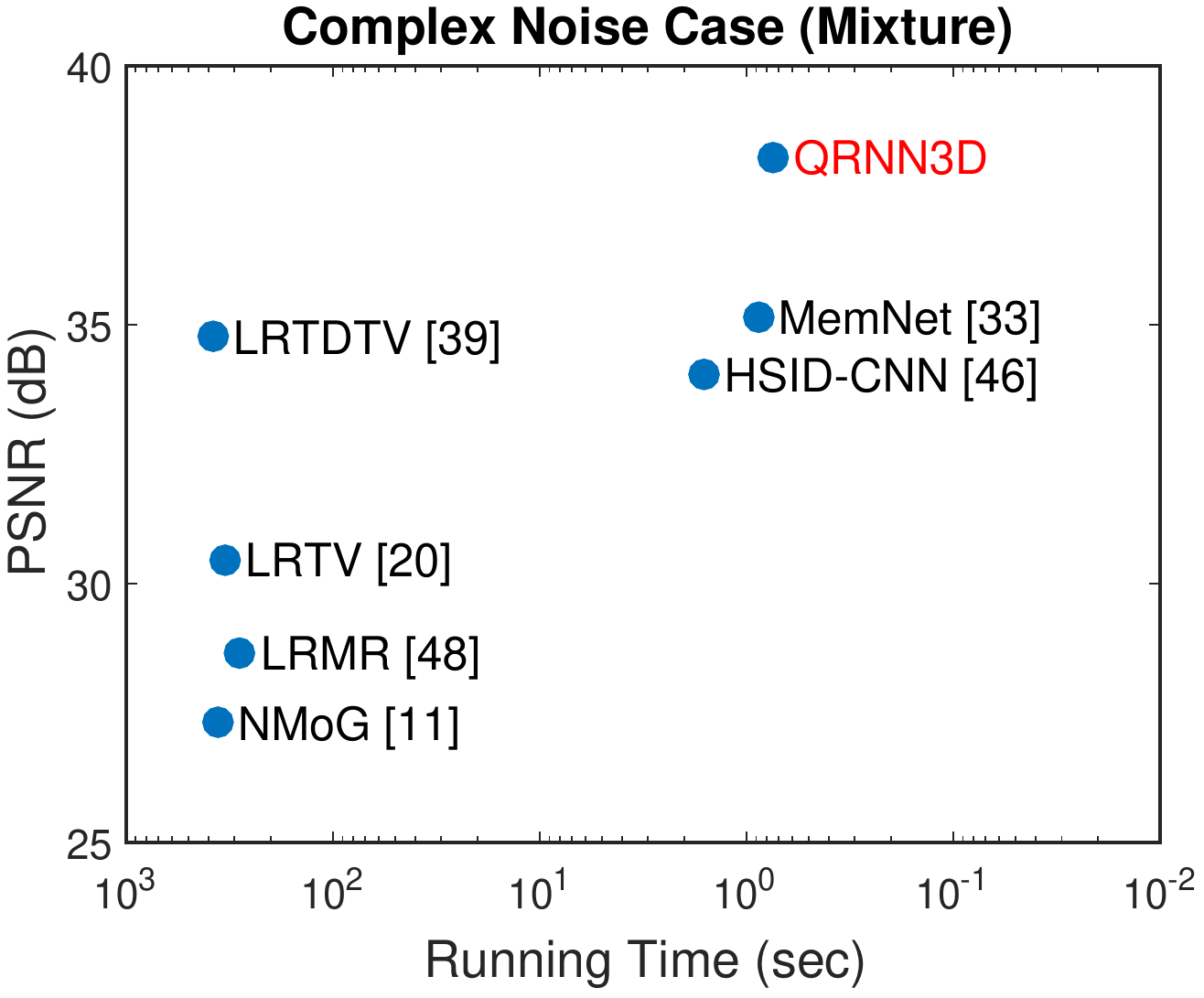}
\label{fig:runtime:2}
\end{subfigure}
\caption{Our QRNN3D outperforms all leading-edge methods on ICVL dataset in both
  Gaussian and complex noise cases. }
\label{fig:runtime}
\end{figure}

Alternative learning-based approaches rely on convolutional neural networks in lieu of the costly optimization and handcrafted priors \cite{8435923,yuan2018hyperspectral}.  Promising results notwithstanding, these approaches model HSI by learned multichannel or band-wise 2D convolutions,
 which sacrifice either the flexibility with respect to the spectral dimension \cite{8435923} (hence requiring retraining network to adapt to HSIs with mismatched spectral dimention), or the model capability to extract GCS knowledge\cite{yuan2018hyperspectral} (thus leading to relatively low performance as shown in Figure \ref{fig:runtime}).

In principal, the trade-off between the model capability and flexibility imposes a fundamental limit for real-world applications. In this paper, we find that combining domain knowledge with 3D deep learning (DL) can achieve both goals
simultaneously. 
Unlike prior DL approaches \cite{8435923,yuan2018hyperspectral} that always utilize  the 2D convolution as a basic building block of network, we introduce a novel building block namely 3D quasi-recurrent
unit (QRU3D) to model HSI from a 3D perspective. This unit contains a 3D convolutional subcomponent and a
quasi-recurrent pooling function \cite{bradbury2017quasi}, enabling structural spatio-spectral
correlation and GCS modeling respectively. The 3D convolutional subcomponent can extract spatio-spectral features from multiple adjacent bands, while the
quasi-recurrent pooling recurrently merges these features over the whole
spectrum, controlled by a dynamic gating mechanism. This mechanism renders the
pooling weights to be dynamically calculated by the input features, thereby
allowing for adaptively modeling the GCS knowledge. 
To eliminate the unidirectional causal dependency (Figure \ref{fig:direction}), introduced by the vanilla recurrent structure, we furthermore propose an alternating directional structure with no additional computation cost. 

Our network, called 3D quasi-recurrent neural network
(QRNN3D), has been designed to make full use of the domain knowledge especially the GCS. It makes significant improvements in model capability/accuracy while is agnostic to the spectral dimension of input HSIs, thus can be applied to any HSIs captured by unknown sensors (with different spectral resolutions). 
 Over extensive experiments, QRNN3D outperforms all leading-edge methods on several benchmark datasets under various noise settings as shown in Figure \ref{fig:runtime}. 

Our main contributions are summarized that we
\begin{enumerate}
\item present a novel building block namely QRU3D that can effectively exploit the domain knowledge -- structural spatio-spectral correlation and global correlation along spectral (GCS) simultaneously. 
\item introduce an alternating directional structure to eliminate the unreasonable causal dependency towards HSI modeling, with no additional computation cost. 
\item demonstrate our model pretrained on ICVL dataset can be directly utilized to tackle remotely sensed imagery which is infeasible in conventional 2D DL approaches for the HSI modeling. 
\end{enumerate}

The remainder of this paper is organized as follows. In Section \ref{sec:rw}, we
review related HSI denoising methods and DL approaches 
that inspire our work. Section \ref{sec:proposed_method} 
introduces the QRNN3D approach for HSI denoising. Extensive experimental results
on natural scenes of HSI database and remote sensed images are presented in
Section \ref{sec:experiment}, followed by more discussions that facilitate the
understanding of QRNN3D in Section \ref{sec:understand}. Conclusions are drawn
in Section \ref{sec:conclusion}.

\section{Related Work} \label{sec:rw}

\subsection{HSI Denoising}
Existing methods towards HSI denoising can be roughly classified into two
categories depending on the noise model.

The most frequently used noise model is zero-mean white and homogeneous Gaussian
additive noise. Under this assumption, BM4D \cite{maggioni2013nonlocal}, an
extension of the BM3D filter \cite{dabov2007BM3D} to volumetric data, could be
directly applied for HSI denoising. By regarding the GCS and non-local
self-similarity in HSI simultaneously, Peng \textit{et al.} proposed a tensor
dictionary learning (TDL) model \cite{peng2014TDL} which achieved very promising
performance. Following this line, more sophisticated methods have been
successively proposed
\cite{dong2015low,fu2017adaptive,zhang2016cluster,chang2017hyper,xie2016multispectral,chang2017weighted, WEI2019412,He_2019_CVPR}.
Among these methods, the low-rank tensor based models, \ie ITS-Reg
\cite{xie2016multispectral}, LLRT \cite{chang2017hyper} and a new iterative projection and denoising algorithm, \ie NG-meet \cite{He_2019_CVPR} achieve
state-of-the-art performance, owing to their elaborate efforts
  on modeling intrinsic property of the HSI. 

  Besides, several works
  \cite{zhang2014hyperspectral,he2016total,xie2016hyperspectral,chen2017denoising,wang2017hyperspectral}
  aim to resolve the realistic complex noise by modeling the noise with
  complicated non-i.i.d. statistical structures. They all frame the denoising
  problem into a low-rank based optimization scheme, and then utilize some
  constraints (\eg total variation, $l_1$ and nuclear norm) to remove the
  complex noise (\eg non-i.i.d. Gaussian, stripe, deadline, impulse).

  Recently, leveraging the power of the DL, Chang \etal \cite{8435923} extended
  the 2D image denoising architecture -- DnCNN \cite{zhang2017beyond} to remove
  various noise in HSIs. They argued the learned filters can well extract the
  structural spatial information. Yuan \etal \cite{yuan2018hyperspectral}
  utilized a deep residual network to recover the remotely
  sensed images under Gaussian noise, which processed HSI with a sliding window
  strategy. Concurrently to our work,  Dong \etal \cite{dong2019deep} proposed a 3D factorizable U-net architecture to exploit spatial-spectral correlations in HSIs from the 3D perspective.  
  All these DL-based methods insufficiently exploit the GCS knowledge, and they
  cannot adjust the learned parameters to adaptively fit input data,
  consequently lacking the freedoms to discriminate the input-dependent
  spatio-spectral correlations.

  In this paper, we leverage the power of the DL to automatically
  learn the mapping purely from the data instead of handcrafted prior and
  complex optimization, reaching to orders-of-magnitude speedup in both
  Gaussian and complex noise contexts. Besides, our DL-based method
  can effectively exploit the underlying characteristics --- structural
  spatio-spectral correlation and GCS, even without sacrificing the flexibility towards HSIs with arbitrary number of bands. 

\subsection{Deep Learning for Image Denoising}
Researches on Gray/RGB image denoising has been dominated by the
discriminative learning based approach especially the deep convolutional neural
network (CNN) in recent years
\cite{zhang2017beyond,mao2016image,tai2017memnet,Chen_2018_ECCV,zhang2018residual,zhang2018RNAN}. Zhang \etal
\cite{zhang2017beyond} proposed a modern deep architecture namely DnCNN by
embedding the batch normalization \cite{ioffe2015batch} and residual learning
\cite{he2016deep}. Meanwhile, Mao \etal \cite{mao2016image} presented a very
deep fully convolutional encoding-decoding framework for image restoration such
as denoising and super-resolution. Both of them yielded better Gaussian denoising
results and less computation time than the highly-engineered benchmark BM3D
\cite{dabov2007BM3D}. Along this line, more works have been proposed to explore
the deep architecture design for image denoising. For example, MemNet \cite{tai2017memnet} introduces memory block to investigate
  the long-term information. 
Residual dense network \cite{zhang2018residual} goes beyond that to build dense connections inner blocks.  
Residual non-local attention network \cite{zhang2018RNAN} utilizes local and non-local attention blocks to extract features that capture the long-range dependencies between pixels and pay more attention to the challenging parts.

Although all these networks can be directly extended into the HSI case, none of
them specifically consider the domain knowledge of the HSI.


\subsection{Deep Image Sequence Modeling}
Modeling image sequence with various lengths is a fundamental problem in a variety
of research fields such as precipitation nowcasting, video processing, and so
on.

Bidirectional recurrent convolutional networks (BRCN)
\cite{huang2015bidirectional} and convolutional LSTM (ConvLSTM)
\cite{xingjian2015convolutional} were proposed for resolving the multi-frame
super-resolution and precipitation nowcasting problem respectively. The key
insight of these models is to replace the common-used recurrent full connections
by weight-sharing convolutional connections such that they can greatly reduce
the large number of network parameters and well model the temporal dependency in
a finer level (i.e. patch-based rather than frame-based). However, these
patch-based operations cannot efficiently capture the spectral correlation, meanwhile recurrently applying convolution along spectrum would drastically increase the
computational complexity. In contrast, our QRNN3D employs an elementwise
recurrent mechanism, enabling good scaling to HSI with a large number of bands. Besides, this mechanism naturally imposes a prior constraint over the spectrum,
making it well-suited for extracting GCS knowledge.

\begin{figure}[!htbp]
\centering
\includegraphics[width=1\linewidth]{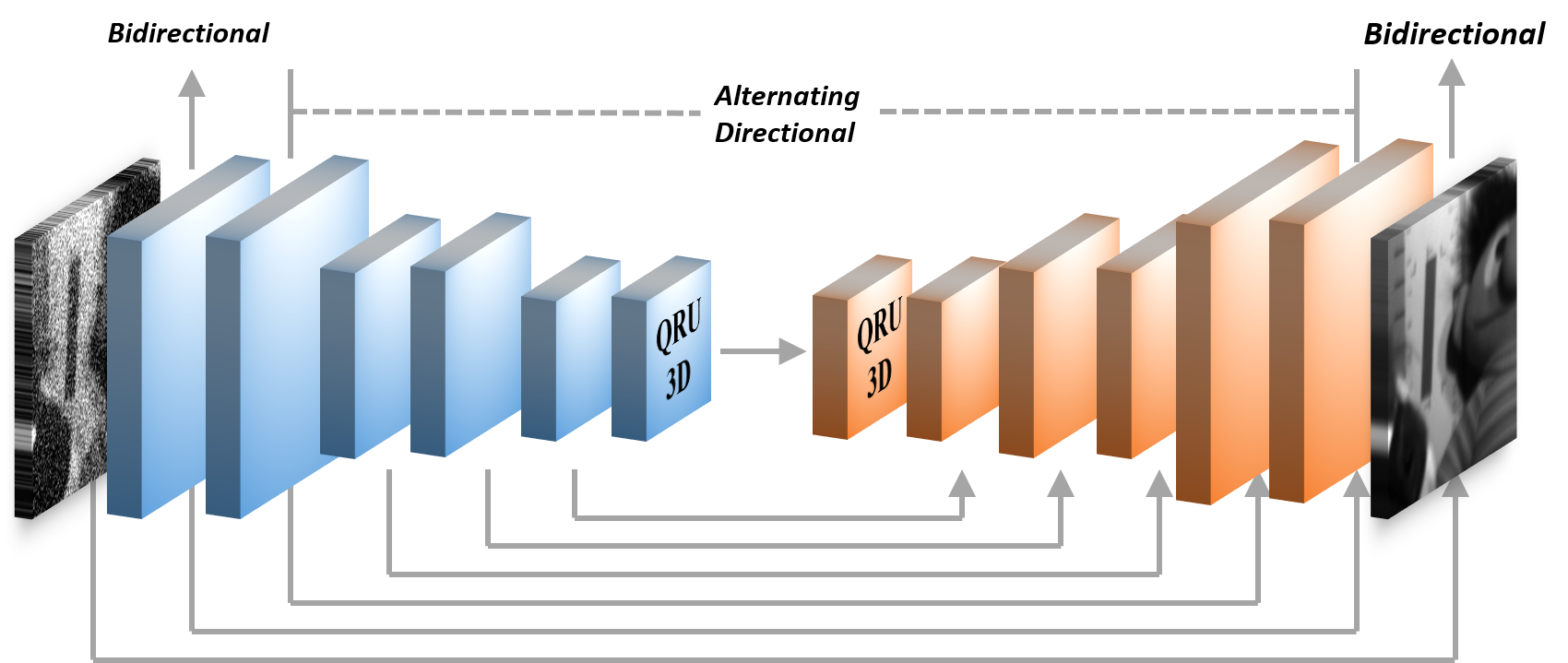}
\caption{The overall architecture of our  residual encoder-decoder QRNN3D. The network
  contains layers of symmetric QRU3D with convolution and deconvolution for
  encoder \textcolor{blue}{(blue)} and decoder \textcolor{orange}{(orange)}
  respectively. Symmetric skip connections are added in each layer. Besides,
  alternating directional structure is equipped in all layers except the top and
  bottom ones with bidirectional structure to avoid bias. }\label{fig:framework}
\end{figure}

\begin{figure*}[!htbp]
\centering
\includegraphics[width=.9\linewidth]{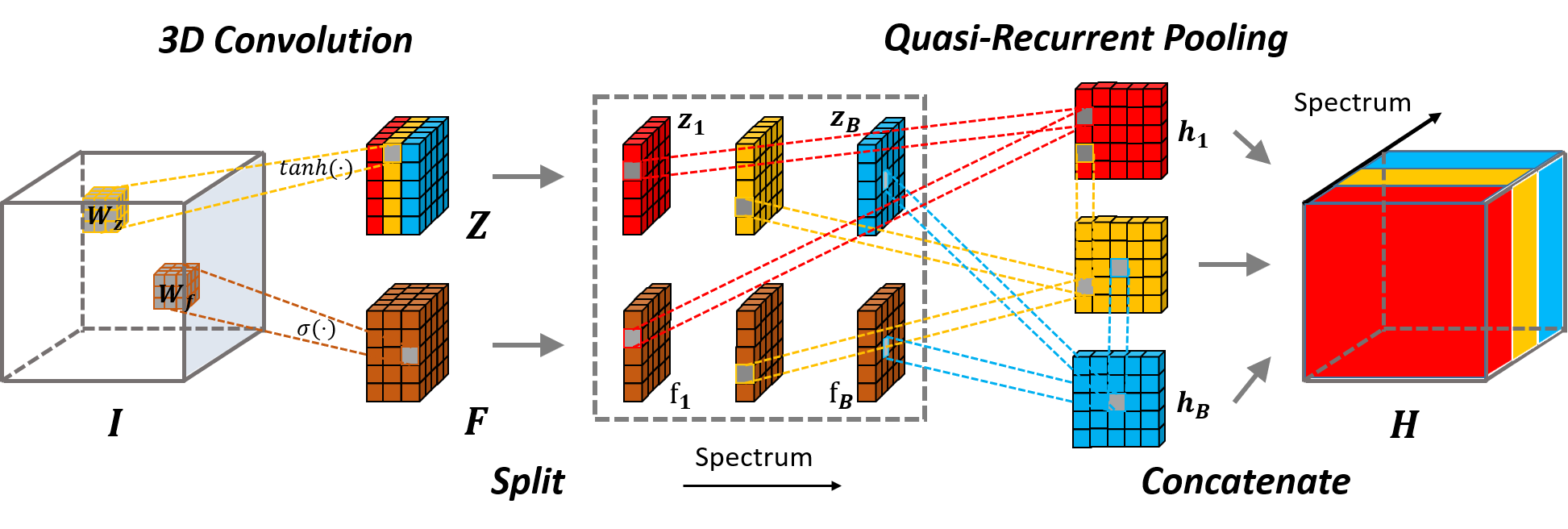}
\caption{The overall structure of QRU3D. It can be
  described in four steps. First, the input $\bm{\mathrm{I}}$ is transformed by
  two set of 3D convolutions, generating a candidate tensor $\bm{\mathrm{Z}}$
  and a neural forget gate $\bm{\mathrm{F}}$. Second, $\bm{\mathrm{Z}}$ and
  $\bm{\mathrm{F}}$ are split along the spectrum to produce sequences of
  $\bm{\mathrm{z}}_b $ and $\bm{\mathrm{f}}_{b}$. Third, the quasi-recurrent
  pooling function is applied recurrently to merge the previous hidden state
  $\bm{\mathrm{h}}_{b-1} $ and current candidate $\bm{\mathrm{z}}_b $ controlled
  by current neural gates $\bm{\mathrm{f}}_{b}$, resulting in a new hidden state
  $\bm{\mathrm{h}}_{b}$. Finally, each hidden state $\bm{\mathrm{h}}_{b}$ is
  concatenated together to form the whole output $\bm{\mathrm{H}}$ to the next
  layer.} \label{fig:qru3d}
\end{figure*}

\begin{table}[!t]
	\centering
	\caption{Network configuration of our residual encoder-decoder style QRNN3D for HSI restoration.}
	\setlength{\tabcolsep}{1mm}{
	\renewcommand{\arraystretch}{1.5}
	\begin{tabular}{c|c|c|c}
	 Layer & $C_{out}$ & Stride & Output size\\ \hline
	 Extractor  & 16 & $1, 1, 1$  & $H\times W \times B$ \\ \hline
	 \multirow{5}{*}{Encoder} 
	 & 16 & $1, 1, 1$ & $H\times W \times B$\\ 
	 & 32 & $2, 2, 1$ & $\frac{H}{2}\times \frac{W}{2} \times B$\\  
	 & 32 & $1, 1, 1$ & $\frac{H}{2}\times \frac{W}{2} \times B$\\ 
	 & 64 & $2, 2, 1$ & $\frac{H}{4}\times \frac{W}{4} \times B$\\ 
	 & 64 & $1, 1, 1$ & $\frac{H}{4} \times \frac{W}{4} \times B$\\ \hline
	 \multirow{5}{*}{Decoder}
	 & 64 & $1, 1, 1$ & $\frac{H}{4} \times \frac{W}{4} \times B$\\  
	 & 32 & $\frac{1}{2}, \frac{1}{2}, 1$ & $\frac{W}{2}\times \frac{W}{2} \times B$\\ 
	 & 32 & $1, 1, 1$ & $\frac{H}{2}\times \frac{W}{2} \times B$\\ 
	 & 16 & $\frac{1}{2}, \frac{1}{2}, 1$ & $H\times W \times B$\\ 
	 & 16 & $1, 1, 1$ & $H \times W \times B$\\ \hline	 
	 Reconstructor  & 1 & $1, 1, 1$  & $H \times W \times B$
	\end{tabular}}
	\label{tb:network}
\end{table}

\section{The Proposed Method} \label{sec:proposed_method}
An HSI degraded by additive noise can be linearly modeled as 
\begin{equation} \label{eq:ObsModel}
\bm{\mathrm{Y}} = \bm{\mathrm{X}}+ \bm{\mathrm{\epsilon}},
\end{equation}
where $\{ \bm{\mathrm{Y}}, \bm{\mathrm{X}}, \bm{\mathrm{\epsilon}} \} \in
\mathbb{R}^{H\times W \times B}$, $\bm{\mathrm{Y}}$ is the observed noisy image,
$\bm{\mathrm{X}}$ is the original clean image, $\bm{\mathrm{\epsilon}}$ denotes
the additive random noise. $H, W, B$ indicate the spatial height, spatial width, 
and number of spectral bands respectively.

Here, we consider miscellaneous noise removal in denoising context, where
$\bm{\mathrm{\epsilon}}$ can represent different types of random noise including
Gaussian noise, sparse noise (stripe, deadline and impulse) or mixture of them.
Given a noisy HSI, our goal is to obtain its noise-free counterpart.

In this section, we introduce the residual encoder-decoder 
QRNN3D for HSI denoising. 
 As shown in Figure
\ref{fig:framework}, our network consists of six pairs of symmetric QRU3D with
convolution and deconvolution for encoder and decoder respectively, leading to
twelve layers in total. We use two layers with stride=2 convolution to
downsample the input in encoder part, and then two layers with stride=1/2 to
upsample in decoder part. The benefits from downsampling and unsampling operations
are that we can use a larger
network under the same computational cost, and increase receptive field size
to make use of the context information in larger image region.
Table \ref{tb:network} illustrates our network configuration. 
Each layer contains a QRU3D with kernel size $3 \times 3 \times 3$, which is set to maximize performance empirically \cite{tran2015learning}. Stride and output
channels ($C_{out}$) in each layer are listed and other
configuration (e.g. padding) can be inferred implicitly.

In the following, we first present the
QRU3D, which is the core building block in our method. Then, alternating
directional structure used to eliminate the unreasonable causal dependency is
introduced, and learning details are provided.

\subsection{3D Quasi-Recurrent Unit}
QRU3D is the basic building block of QRNN3D. It consists of two subcomponents,
\ie 3D convolutional subcomponent and quasi-recurrent pooling, as shown in
Figure \ref{fig:qru3d}. Unlike the 2D convolution, both of the subcomponents do not
enforce the number of 
spectral bands, making the QRNN3D free for
processing HSIs with arbitrary bands.

\subsubsection{3D Convolutional Subcomponent}
The 3D convolutional subcomponent of QRU3D performs two set of 3D convolutions
\cite{ji20133d,tran2015learning} with separated filter banks, producing sequence
of tensors passed through different activation functions,
\begin{align}
\begin{split} 
  \bm{\mathrm{Z}} &= \mathop{\mathrm{tanh}}(\bm{\mathrm{W}}_z * \bm{\mathrm{I}}), \\
  \bm{\mathrm{F}} &= \mathop{\mathrm{\sigma}}(\bm{\mathrm{W}}_{f} *
  \bm{\mathrm{I}}),
\end{split} 
\end{align}
where $\bm{\mathrm{I}} \in \mathbb{R}^{C_{in} \times H\times W \times B}$ is the
input feature maps coming from last layer (in first layer, input
$\bm{\mathrm{I}} = \bm{\mathrm{Y}}$ with $C_{in} = 1$); $\bm{\mathrm{Z}} \in
\mathbb{R}^{C_{out}\times H\times W \times B}$ is a high dimensional candidate
tensor. $\bm{\mathrm{F}}$ has the same dimension as $\bm{\mathrm{Z}}$,
representing the neural forget gate that controls the behavior of dynamic
memorization. Both $\bm{\mathrm{W}}_z$ and $\bm{\mathrm{W}}_{f} \in
\mathbb{R}^{C_{out} \times C_{in}\times 3\times 3 \times 3}$ are the 3D
convolutional filter banks and $*$ denotes a 3D convolution, $\sigma$ indicates
a sigmoid non-linearity.

The 3D convolution  is achieved by convolving a 3D kernel to a whole HSI in
both spatial and spectral dimensions. 
The 3D convolution in the spatial domain can mimic numerous operations widely
used in low-level vision (like image patch extraction and 2D patch transform in
BM3D \cite{dabov2007BM3D,Lefkimmiatis_2017_CVPR}) and the 3D convolution in the
spectral domain can model the local spectrum continuity to alleviate the
spectral distortion. Consequently, the embedded C3D can effectively exploit the
structural spatio-spectral correlation in HSIs.



\subsubsection{Quasi-Recurrent Pooling}
Although the 3D convolutional subcomponent has already exploited the inter-band
relationship, it is computed in a local way and cannot explicitly exploit GCS. To
effectively utilize the GCS, we present quasi-recurrent pooling, in which
pooling operation and dynamic gating mechanism are introduced.


In our QRU3D, the quasi-recurrent pooling is applied after the candidate tensor
$\bm{\mathrm{Z}}$ and neural forget gate $\bm{\mathrm{F}}$ are obtained by the
3D convolutional subcomponent. We first split $\bm{\mathrm{Z}}$ and
$\bm{\mathrm{F}}$ along the spectrum, generating sequences of  $\bm{\mathrm{z}}_b
$ and $\bm{\mathrm{f}}_{b}$ respectively, and then feed these states 
into a
quasi-recurrent pooling function \cite{bradbury2017quasi}, 
\begin{equation} 
\bm{\mathrm{h}}_b = \bm{\mathrm{f}}_b \odot \bm{\mathrm{h}}_{b-1} + (1-\bm{\mathrm{f}}_b) \odot \bm{\mathrm{z}}_b, \quad  \forall b \in [1, B] ,
\label{eq:f}
\end{equation} 
where $\odot$ denotes an element-wise multiplication, 
$\bm{\mathrm{h}}_{b-1}$ is the
hidden state merged through all previous states and also represents the $(b-1)$-th
band in the output of this layer,
$\bm{\mathrm{h}}_0 = \bf 0$ with all entries
equal to zero. The forget gate $\bm{\mathrm{f}}_b$ balances the weight of
current candidate $\bm{\mathrm{z}}_b $ and previous memory, \ie hidden state
$\bm{\mathrm{h}}_{b-1}$.
Its value depends on the current input $\bm{\mathrm{I}}$ instead of being fixed
like a convolutional filter, which can effectively adapt to the input image own
and not solely rely on the parameters learned in the training stage. By this
construction, the inter-band information would be accurately merged. Meanwhile,
since this dynamic pooling recurrently operates across the whole spectrum, the
GCS can be effectively exploited. The output feature maps $\bm{\mathrm{H}}$ will
be produced by concatenating all hidden states along the spectrum.

In addition, due to independent neural gate and element-wise recurrent
operations (multiplication), the QRU3D is highly parallel, enabling good scaling
to HSI with a large number of bands. More specifically, the calculation of
neural forget gate $\bm{\mathrm{f}}_b$ is only dependent on multiple contiguous
bands of input instead of involving the previous hidden state in typical RNNs
(\eg LSTM \cite{hochreiter1997long} and GRU \cite{cho2014learning}). Meanwhile,
the elementwise multiplication is exceedingly computationally economical than
the convolution used by ConvLSTM \cite{xingjian2015convolutional}, thus can be
easily recurrently utilized hundreds of times.

\begin{figure}[!htbp]
\centering
\begin{subfigure}[b]{.3\linewidth}
\centering
\includegraphics[width=1\linewidth]{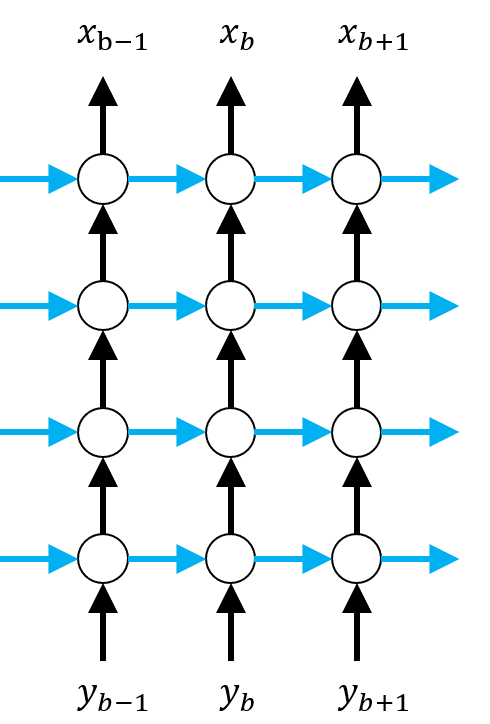}

\caption{}
\label{fig:unidirection}
\end{subfigure}
\begin{subfigure}[b]{.3\linewidth}
\centering
\includegraphics[width=1\linewidth]{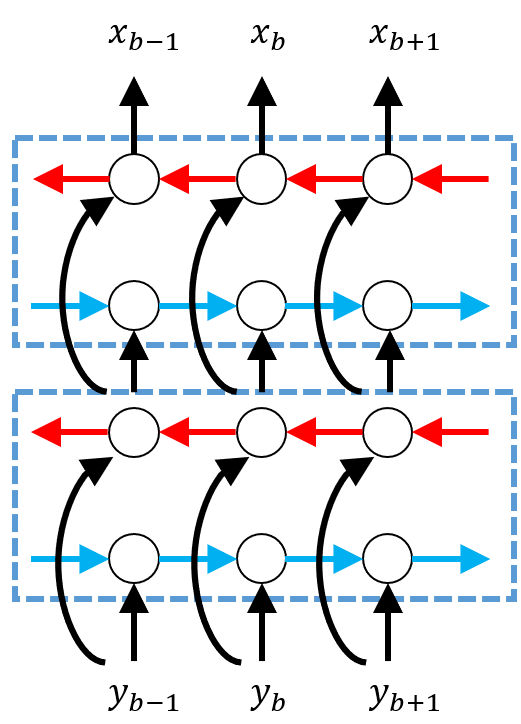}

\caption{}
\label{fig:bidirection}
\end{subfigure}
\begin{subfigure}[b]{.3\linewidth}
\centering
\includegraphics[width=1\linewidth]{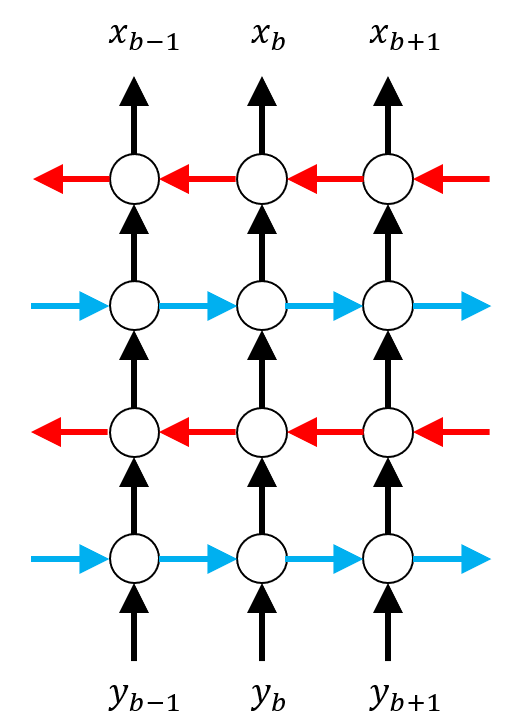}

\caption{}
\label{fig:alterdirection}
\end{subfigure}

\caption{Directional structure overview. (a) Unidirectional structure: hidden
  states propagate unidirectionally. (b) Bidirectional structure: one layer
  contains two sublayers which propagate states with inverse direction,
  generating results by adding sublayers' output. (c) Our proposed alternating
  directional structure: direction of network changes in each
  layer.}\label{fig:direction}
\end{figure}

\begin{figure}[!t]
\centering
\setlength\tabcolsep{0pt}
\renewcommand\arraystretch{0}
\begin{tabular}{ccccccc}
\includegraphics[height=.13\linewidth,clip,keepaspectratio]{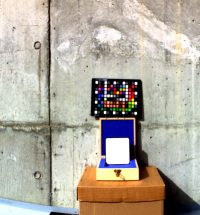} &
\includegraphics[height=.13\linewidth,clip,keepaspectratio]{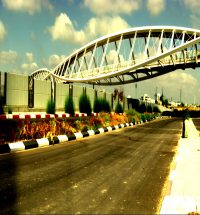} &
\includegraphics[height=.13\linewidth,clip,keepaspectratio]{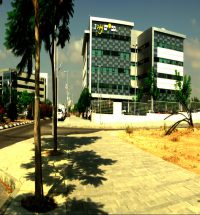} &
\includegraphics[height=.13\linewidth,clip,keepaspectratio]{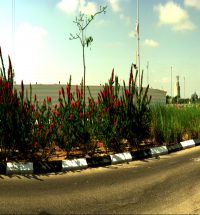} &
\includegraphics[height=.13\linewidth,clip,keepaspectratio]{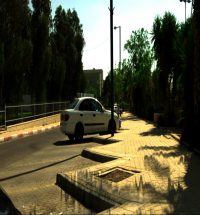} &
\includegraphics[height=.13\linewidth,clip,keepaspectratio]{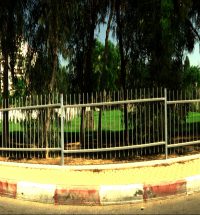} &
\includegraphics[height=.13\linewidth,clip,keepaspectratio]{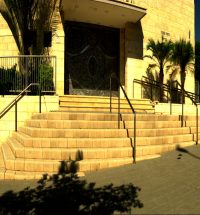} \\

\includegraphics[height=.13\linewidth,clip,keepaspectratio]{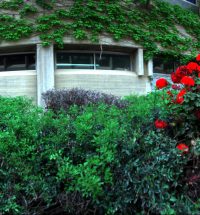} &
\includegraphics[height=.13\linewidth,clip,keepaspectratio]{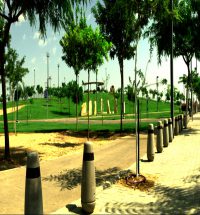} &
\includegraphics[height=.13\linewidth,clip,keepaspectratio]{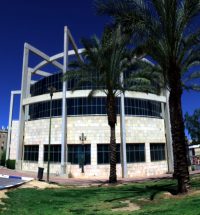} &
\includegraphics[height=.13\linewidth,clip,keepaspectratio]{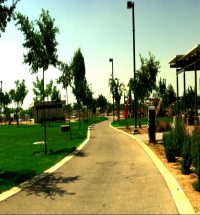} &
\includegraphics[height=.13\linewidth,clip,keepaspectratio]{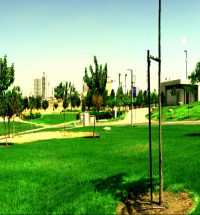} &
\includegraphics[height=.13\linewidth,clip,keepaspectratio]{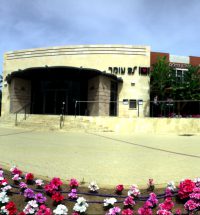} &
\includegraphics[height=.13\linewidth,clip,keepaspectratio]{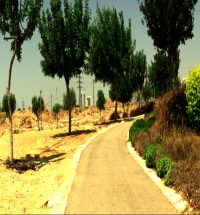} \\

\includegraphics[height=.13\linewidth,clip,keepaspectratio]{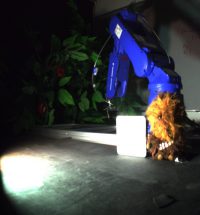} &
\includegraphics[height=.13\linewidth,clip,keepaspectratio]{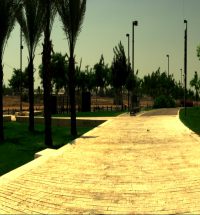} &
\includegraphics[height=.13\linewidth,clip,keepaspectratio]{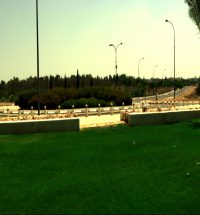} &
\includegraphics[height=.13\linewidth,clip,keepaspectratio]{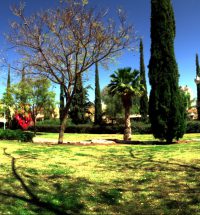} &
\includegraphics[height=.13\linewidth,clip,keepaspectratio]{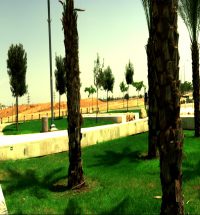} &
\includegraphics[height=.13\linewidth,clip,keepaspectratio]{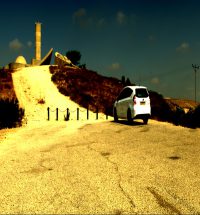} &
\includegraphics[height=.13\linewidth,clip,keepaspectratio]{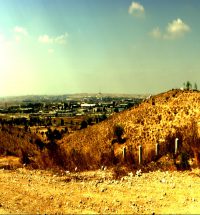} \\

\includegraphics[height=.13\linewidth,clip,keepaspectratio]{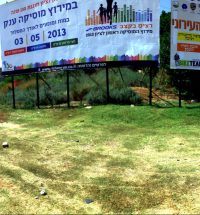} &
\includegraphics[height=.13\linewidth,clip,keepaspectratio]{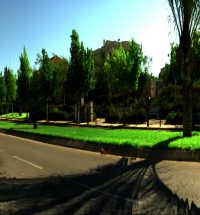} &
\includegraphics[height=.13\linewidth,clip,keepaspectratio]{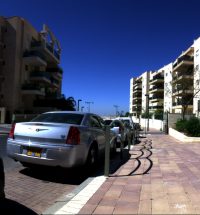} &
\includegraphics[height=.13\linewidth,clip,keepaspectratio]{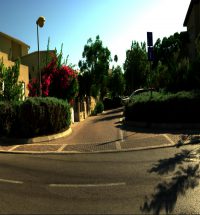} &
\includegraphics[height=.13\linewidth,clip,keepaspectratio]{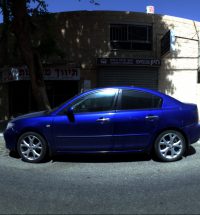} &
\includegraphics[height=.13\linewidth,clip,keepaspectratio]{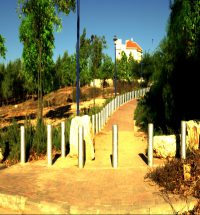} &
\includegraphics[height=.13\linewidth,clip,keepaspectratio]{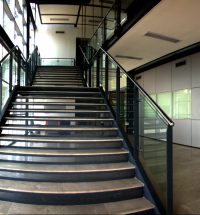} 

\end{tabular}
\caption{Synthesized RGB image samples from ICVL dataset.}
\label{fig:icvl_rgb}
\end{figure}

\subsection{Alternating Directional Structure}
A forward 3D quasi-recurrent unit, as in Equation \eqref{eq:f}, reads a
candidate tensor 
$\bm{\mathrm{z}}_b$ in
order starting from the first $\bm{\mathrm{z}}_1$ to the last
$\bm{\mathrm{z}}_B$, so that a hidden state $\bm{\mathrm{h}}_b$ only depends on
the previous $\bm{\mathrm{z}}_b$ (and theirs corresponding bands). This
introduces the causal dependency since the computing stream of hidden state
propagates unidirectionally as shown in Figure \ref{fig:unidirection}, which is
not reasonable for the HSI. 

A typical solution is to use a bidirectional
structure
\cite{schuster1997bidirectional,huang2015bidirectional,bahdanau2015neural}, in
which a layer of network contains two sublayers, \ie a forward QRU3D and a
backward QRU3D in our case, as shown in Figure \ref{fig:bidirection}. The forward
QRU3D reads the candidate tensor sequence in order and calculates a sequence of
forward hidden states. The backward QRU3D reads the sequence in reverse
order, leading to a sequence of backward hidden states. The output of
this layer is calculated by adding the forward and backward hidden states
elementwisely. However, this structure makes the computational burden
unacceptable because of the nearly double amount of memory consumption.

 To ease this issue, we present an alternating directional structure for HSIs. In
specific, a QRNN3D with alternating directional structure changes the direction
of computing stream of hidden state in each layer, as shown in Figure
\ref{fig:alterdirection}. This structure is built by alternately stacking
forward and backward QRU3D, in which a forward (or backward) state 
is be merged by a backward (or
forward) state in next layer, such that the global context information could be
propagated through the whole spectrum.

Compared with the typical solution by bidirectional structure, our proposed
alternating directional structure almost adds no additional computation cost, while
keeping the ability to model the dependency from whole spectrum of an HSI
regardless of the position of the output.

\begin{table*}[!htbp]
	\centering
	\caption{Overview of our incremental train policy. Our network learning
goes through three stages, from the easy task of Gaussian
denoising with fixed noise level, to the difficult one of complex
noise removal.  In our implementation, fixed noise level $\sigma$ in stage 1 is set to 50. Unknown $\sigma$ in stage 2 is uniformly sampled from 30 to 70. Unknown complex noise in stage 3 denotes the complex noise randomly chosen from Case 1 to 4 (without Case 5: mixture noise). 
The models trained at the end of stage 2 (epoch 50) and 3 (epoch 100) are used in Gaussian denoising and complex noise removal tasks respectively.}
	\setlength{\tabcolsep}{1mm}{
	\begin{tabular}{|C{.1\linewidth}|C{.1\linewidth}|C{.1\linewidth}|C{.1\linewidth}|C{.1\linewidth}|C{.1\linewidth}|C{.1\linewidth}|C{.1\linewidth}|C{.1\linewidth}|}
		\hline
		Stage &  \multicolumn{2}{c|}{1} &  \multicolumn{3}{c|}{2} & \multicolumn{3}{c|}{3}  \\ \hline
		Noise model &  \multicolumn{2}{c|}{Gaussian noise with known $\sigma$} & \multicolumn{3}{c|}{Gaussian noise with unknown $\sigma$} & \multicolumn{3}{c|}{Unknown complex noise}  \\ \hline
		Epoch & 0 $\sim$ 20 & 20 $\sim$ 30 & 30 $\sim$ 35 & 35 $\sim$ 45 & 45 $\sim$ 50 & 50 $\sim$ 85 & 85 $\sim$ 95 & 95 $\sim$ 100  \\ \hline
		Learning rate & $10^{-3}$ & $10^{-4}$ & $10^{-3}$ & $10^{-4}$ & $10^{-5}$ & $10^{-3}$ & $10^{-4}$ & $10^{-5}$  \\ \hline
		Batch size &  \multicolumn{2}{c|}{16} &  \multicolumn{6}{c|}{64} \\  \hline
	\end{tabular}}
	\label{tb:training-policy}
\end{table*}

\section{Experimental Results} \label{sec:experiment}

\begin{figure*}[!htbp]
\centering
\begin{subfigure}[b]{.116\linewidth}
\centering
\includegraphics[width=1\linewidth,clip,keepaspectratio]{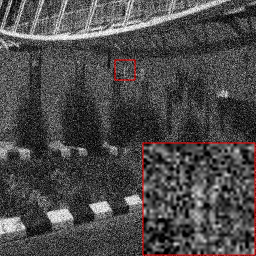}
\caption{Noisy \protect\\ \centering (14.17)}
\end{subfigure}
\begin{subfigure}[b]{.116\linewidth}
\centering
\includegraphics[width=1\linewidth,clip,keepaspectratio]{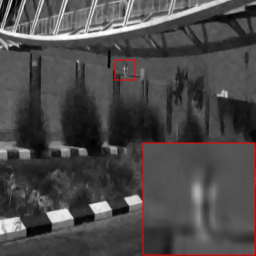}
\caption{BM4D \protect\\ \centering (33.00)}
\end{subfigure}
\begin{subfigure}[b]{.116\linewidth}
\centering
\includegraphics[width=1\linewidth,clip,keepaspectratio]{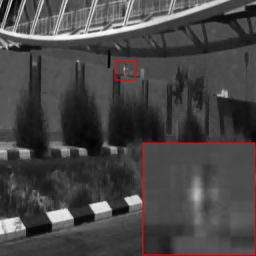}
\caption{TDL \protect\\ \centering (35.11)}
\end{subfigure}
\begin{subfigure}[b]{.116\linewidth}
\centering
\includegraphics[width=1\linewidth,clip,keepaspectratio]{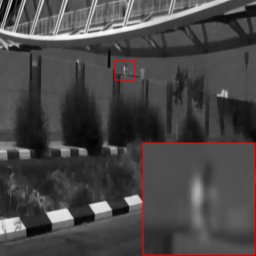}
\caption{ITSReg \protect\\ \centering (36.09)}
\end{subfigure}
\begin{subfigure}[b]{.116\linewidth}
\centering
\includegraphics[width=1\linewidth,clip,keepaspectratio]{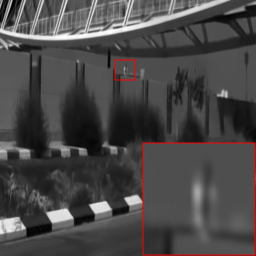}
\caption{LLRT \protect\\ \centering (36.08)}
\end{subfigure}
\begin{subfigure}[b]{.116\linewidth}
\centering
\includegraphics[width=1\linewidth,clip,keepaspectratio]{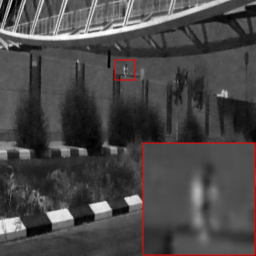}
\caption{HSID-CNN \protect\\ \centering (35.22)}
\end{subfigure}
\begin{subfigure}[b]{.116\linewidth}
\centering
\includegraphics[width=1\linewidth,clip,keepaspectratio]{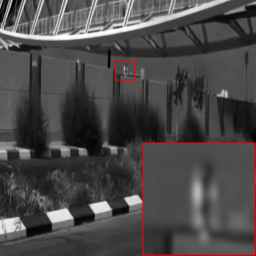}
\caption{MemNet \protect\\ \centering (36.29)}
\end{subfigure}
\begin{subfigure}[b]{.116\linewidth}
\centering
\includegraphics[width=1\linewidth,clip,keepaspectratio]{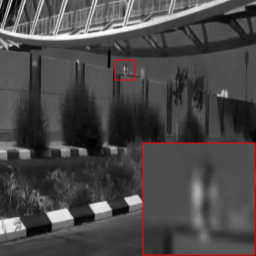}
\caption{Ours \protect\\ \centering (36.73)}
\end{subfigure}

\caption{Simulated Gaussian noise removal results of PSNR (dB) at $20^{th}$ band of image under noise level $\sigma=50 $ on ICVL dataset. (\textbf{Best view on screen with zoom})}
\label{fig:denoising:1} 
\end{figure*}

\begin{figure*}[!t]
	\centering
	\setlength\tabcolsep{1pt}
	\renewcommand\arraystretch{1}
	\begin{tabular}{ccccccccc}
	& Noisy & LRMR\cite{zhang2014hyperspectral} & LRTV\cite{he2016total} & NMoG\cite{chen2017denoising} & TDTV\cite{wang2017hyperspectral} & D-CNN\cite{yuan2018hyperspectral} & MemNet\cite{tai2017memnet} & Ours \\
\specialrule{1pt}{3pt}{4pt}
\rotatebox[origin=c]{90}{Case 1}
& \includegraphics[align=c,width=.116\linewidth,clip,keepaspectratio]{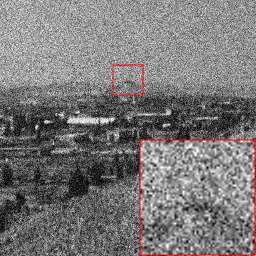}
& \includegraphics[align=c,width=.116\linewidth,clip,keepaspectratio]{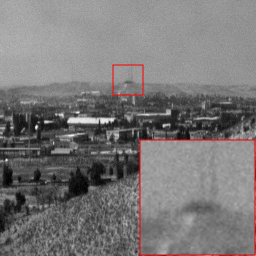}
& \includegraphics[align=c,width=.116\linewidth,clip,keepaspectratio]{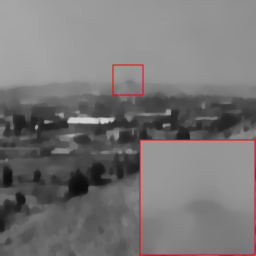}
& \includegraphics[align=c,width=.116\linewidth,clip,keepaspectratio]{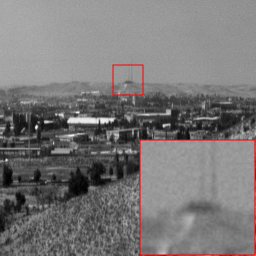}
& \includegraphics[align=c,width=.116\linewidth,clip,keepaspectratio]{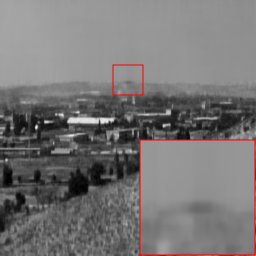}
& \includegraphics[align=c,width=.116\linewidth,clip,keepaspectratio]{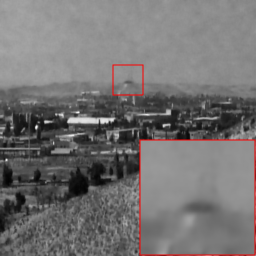}
& \includegraphics[align=c,width=.116\linewidth,clip,keepaspectratio]{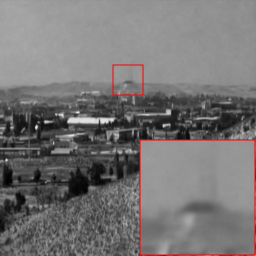}
& \includegraphics[align=c,width=.116\linewidth,clip,keepaspectratio]{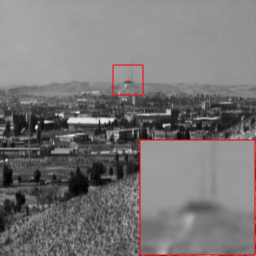} \\
\specialrule{1pt}{3pt}{4pt}
\rotatebox[origin=c]{90}{Case 2}
& \includegraphics[align=c,width=.116\linewidth,clip,keepaspectratio]{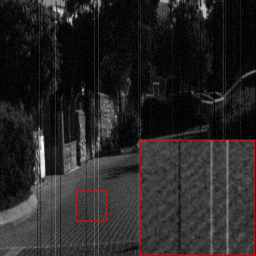}
& \includegraphics[align=c,width=.116\linewidth,clip,keepaspectratio]{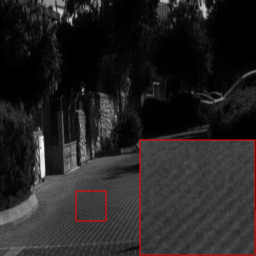}
& \includegraphics[align=c,width=.116\linewidth,clip,keepaspectratio]{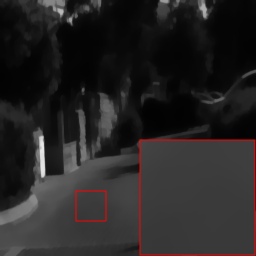}
& \includegraphics[align=c,width=.116\linewidth,clip,keepaspectratio]{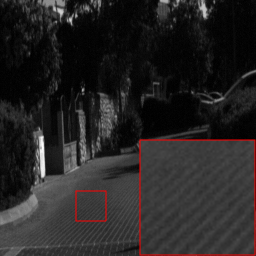}
& \includegraphics[align=c,width=.116\linewidth,clip,keepaspectratio]{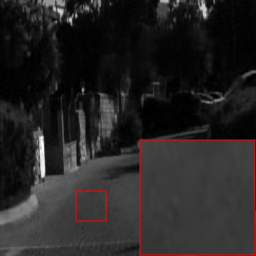}
& \includegraphics[align=c,width=.116\linewidth,clip,keepaspectratio]{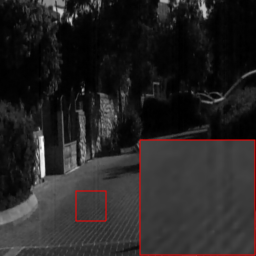}
& \includegraphics[align=c,width=.116\linewidth,clip,keepaspectratio]{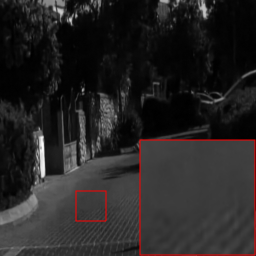}
& \includegraphics[align=c,width=.116\linewidth,clip,keepaspectratio]{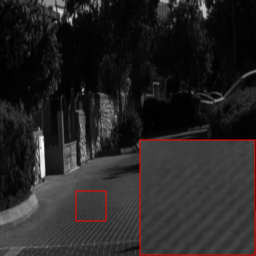} \\
\specialrule{1pt}{3pt}{4pt}
\rotatebox[origin=c]{90}{Case 3}
& \includegraphics[align=c,width=.116\linewidth,clip,keepaspectratio]{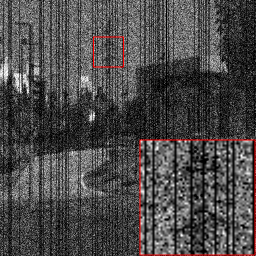}
& \includegraphics[align=c,width=.116\linewidth,clip,keepaspectratio]{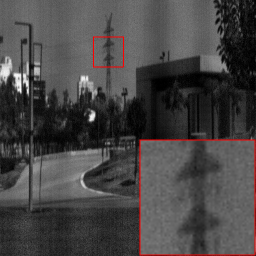}
& \includegraphics[align=c,width=.116\linewidth,clip,keepaspectratio]{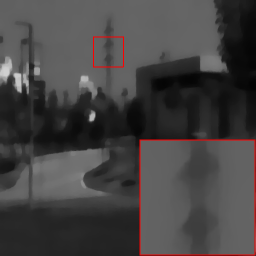}
& \includegraphics[align=c,width=.116\linewidth,clip,keepaspectratio]{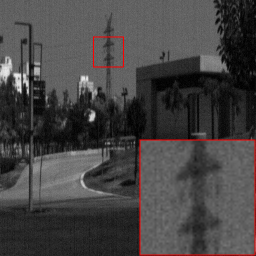}
& \includegraphics[align=c,width=.116\linewidth,clip,keepaspectratio]{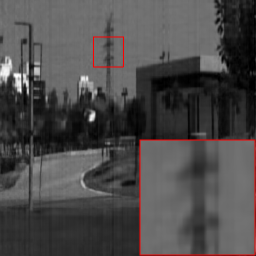}
& \includegraphics[align=c,width=.116\linewidth,clip,keepaspectratio]{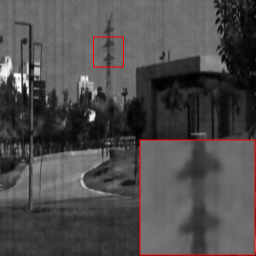}
& \includegraphics[align=c,width=.116\linewidth,clip,keepaspectratio]{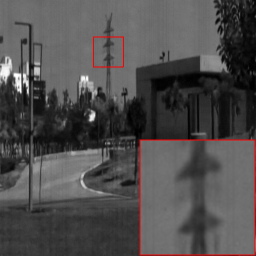}
& \includegraphics[align=c,width=.116\linewidth,clip,keepaspectratio]{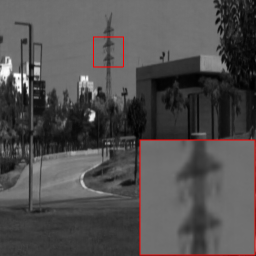} \\
\specialrule{1pt}{3pt}{4pt}
\rotatebox[origin=c]{90}{Case 4}
& \includegraphics[align=c,width=.116\linewidth,clip,keepaspectratio]{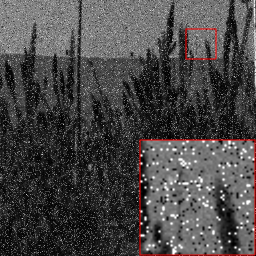}
& \includegraphics[align=c,width=.116\linewidth,clip,keepaspectratio]{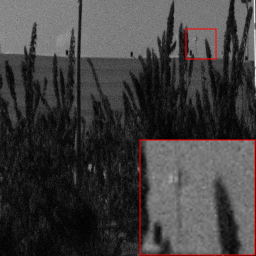}
& \includegraphics[align=c,width=.116\linewidth,clip,keepaspectratio]{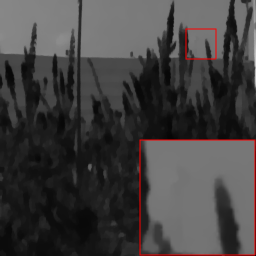}
& \includegraphics[align=c,width=.116\linewidth,clip,keepaspectratio]{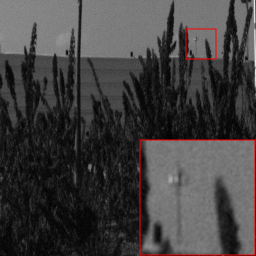}
& \includegraphics[align=c,width=.116\linewidth,clip,keepaspectratio]{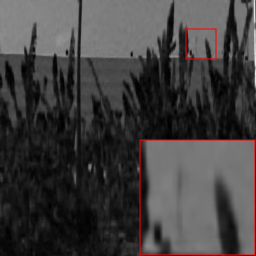}
& \includegraphics[align=c,width=.116\linewidth,clip,keepaspectratio]{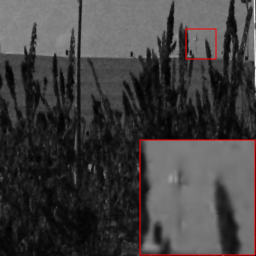}
& \includegraphics[align=c,width=.116\linewidth,clip,keepaspectratio]{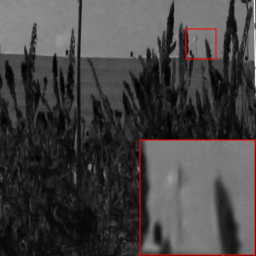}
& \includegraphics[align=c,width=.116\linewidth,clip,keepaspectratio]{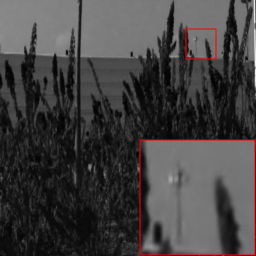} \\
\specialrule{1pt}{3pt}{4pt}
\rotatebox[origin=c]{90}{Case 5}
& \includegraphics[align=c,width=.116\linewidth,clip,keepaspectratio]{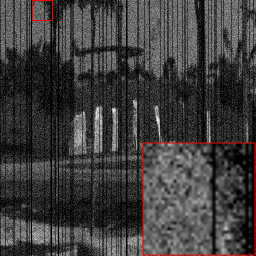}
& \includegraphics[align=c,width=.116\linewidth,clip,keepaspectratio]{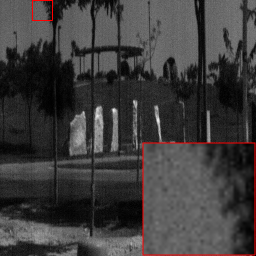}
& \includegraphics[align=c,width=.116\linewidth,clip,keepaspectratio]{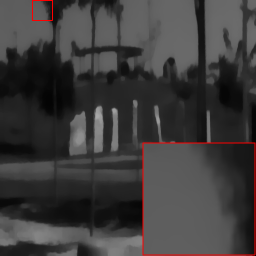}
& \includegraphics[align=c,width=.116\linewidth,clip,keepaspectratio]{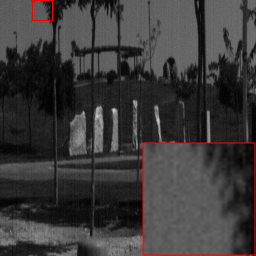}
& \includegraphics[align=c,width=.116\linewidth,clip,keepaspectratio]{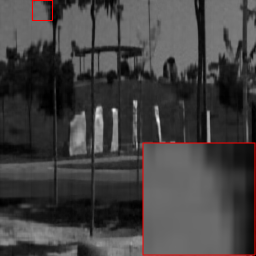}
& \includegraphics[align=c,width=.116\linewidth,clip,keepaspectratio]{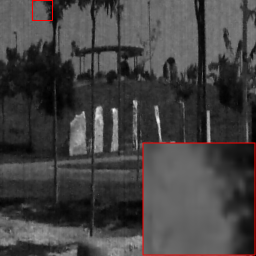}
& \includegraphics[align=c,width=.116\linewidth,clip,keepaspectratio]{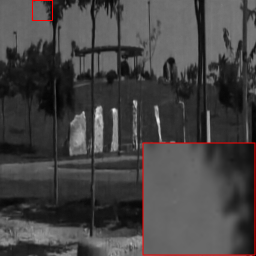}
& \includegraphics[align=c,width=.116\linewidth,clip,keepaspectratio]{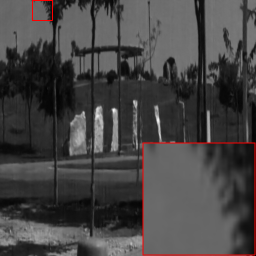} \\
\specialrule{1pt}{3pt}{4pt}
\end{tabular}
\caption{Simulated complex noise removal result s on ICVL dataset. Examples for non-i.i.d Gaussian noise, Gaussian + stripes, Gaussian + deadline, Gaussian + impulse and mixture noise removal (Cases 1-5) are presented respectively. (\textbf{Best view on screen with zoom})}
\label{fig:denoising:icvl_complex}
\end{figure*}

\begin{figure*}
\centering
\begin{subfigure}[b]{.3\textwidth}
\centering
\includegraphics[width=1\linewidth,clip,keepaspectratio]{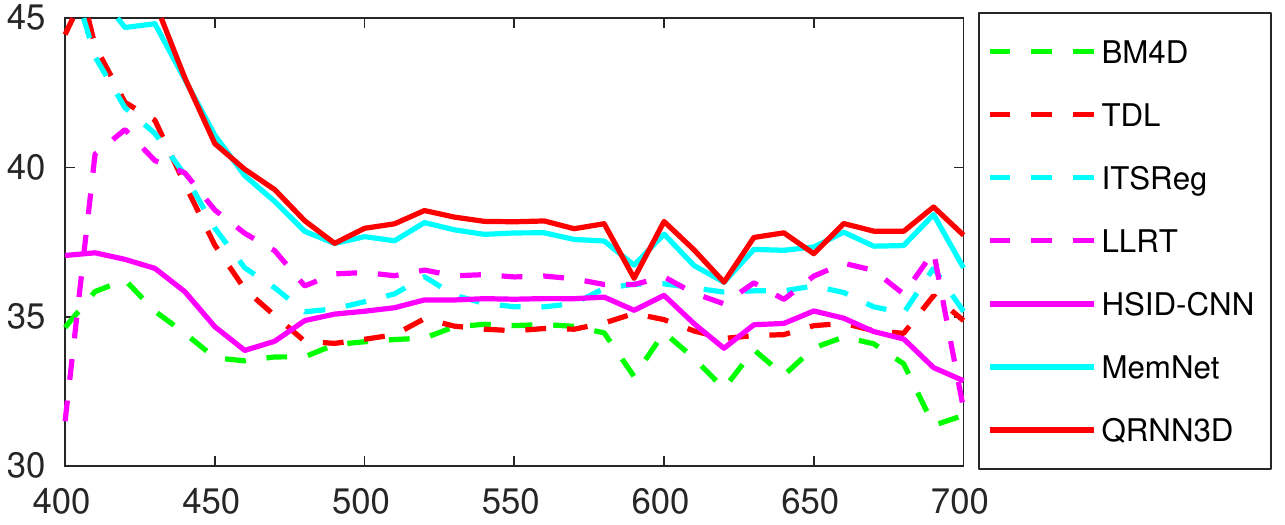}
\caption{i.i.d. Gaussian ($\sigma = 50)$}
\end{subfigure}
\begin{subfigure}[b]{.3\textwidth}
\centering
\includegraphics[width=1\linewidth,clip,keepaspectratio]{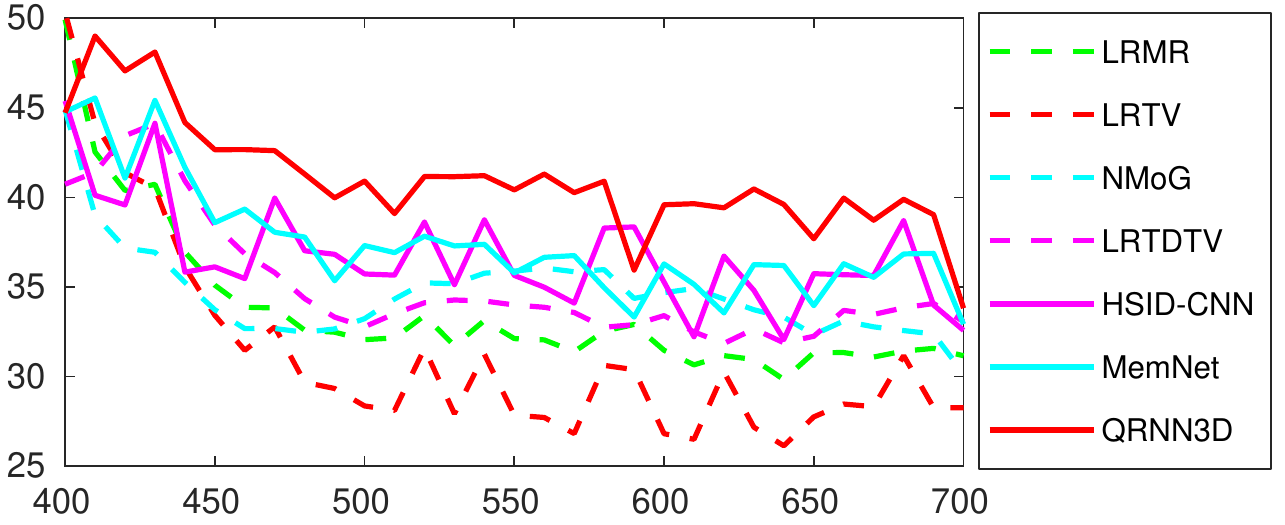}
\caption{Non-i.i.d. Gaussian (Case 1)}
\end{subfigure}
\begin{subfigure}[b]{.3\textwidth}
\centering
\includegraphics[width=1\linewidth,clip,keepaspectratio]{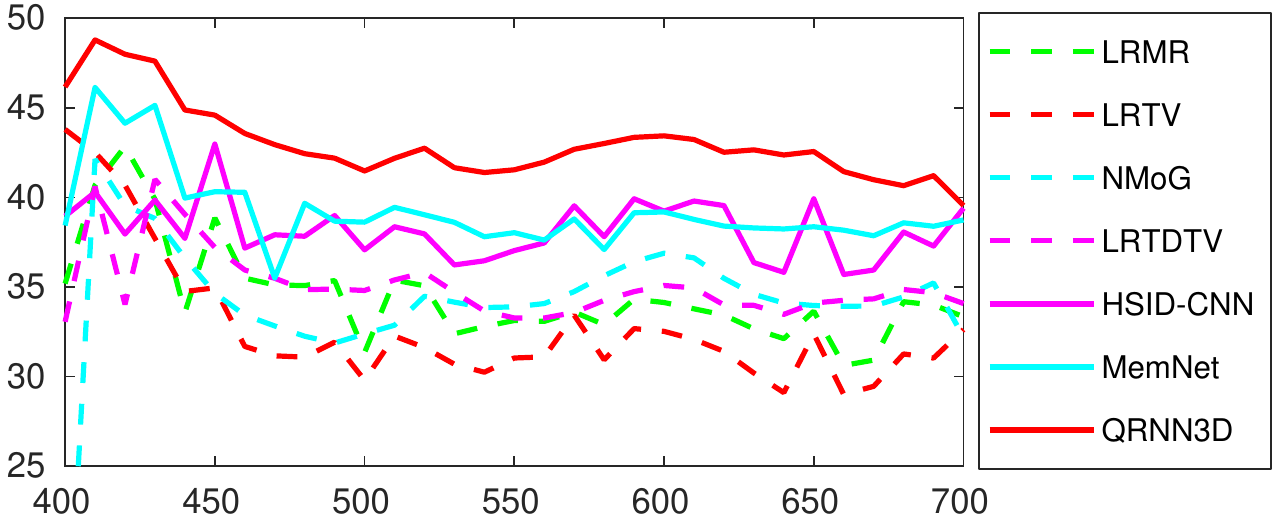}
\caption{Gaussian + Stripe (Case 2)}
\end{subfigure}
\begin{subfigure}[b]{.3\textwidth}
\centering
\includegraphics[width=1\linewidth,clip,keepaspectratio]{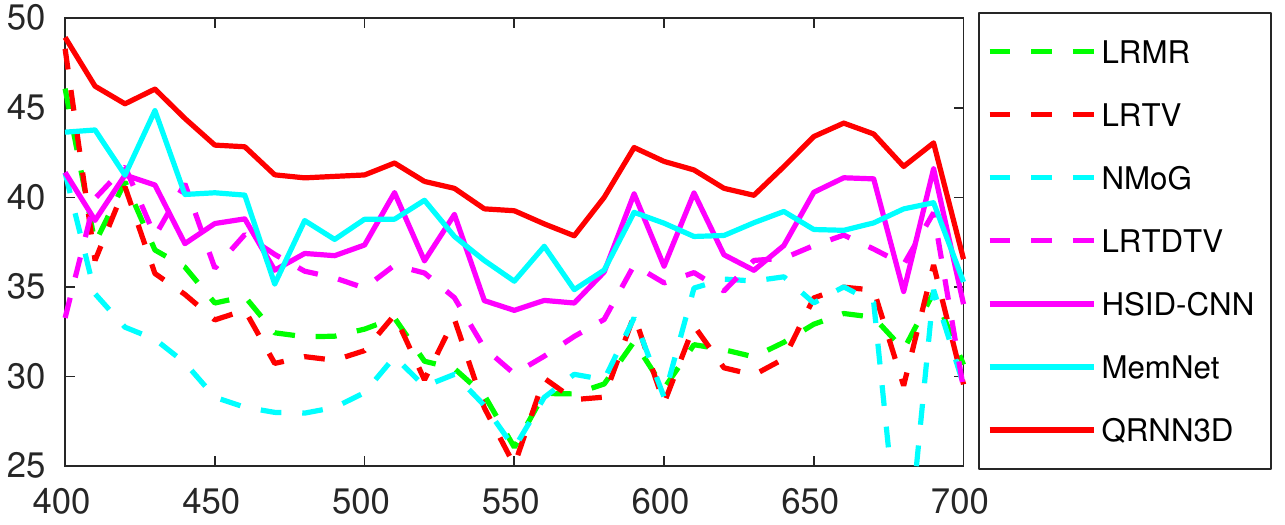}
\caption{Gaussian + Deadline (Case 3)}
\end{subfigure}
\begin{subfigure}[b]{.3\textwidth}
\centering
\includegraphics[width=1\linewidth,clip,keepaspectratio]{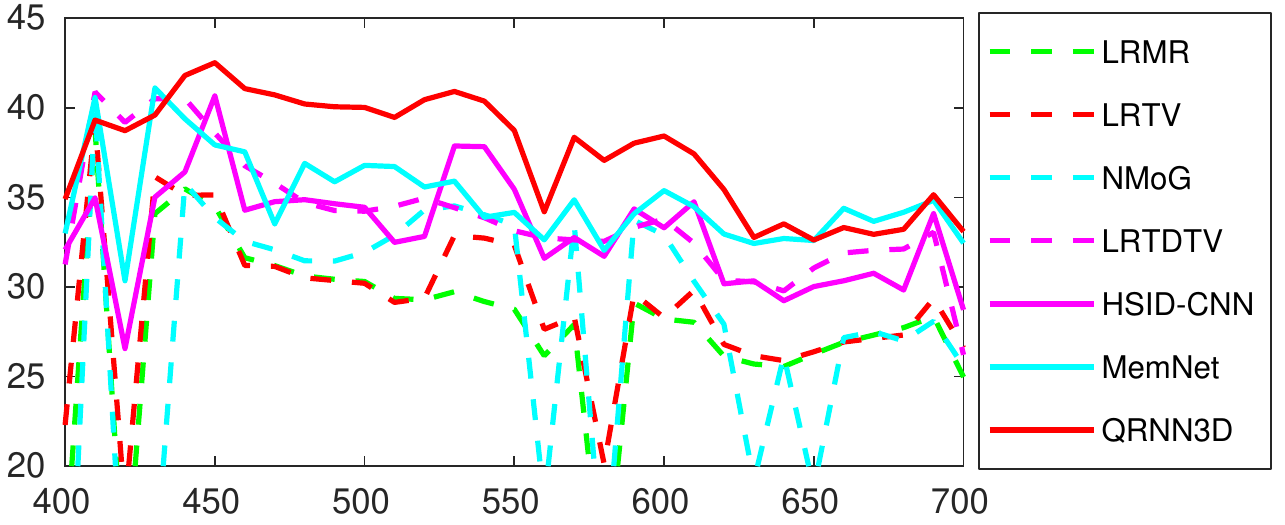}
\caption{Gaussian + Impulse (Case 4)}
\end{subfigure}
\begin{subfigure}[b]{.3\textwidth}
\centering
\includegraphics[width=1\linewidth,clip,keepaspectratio]{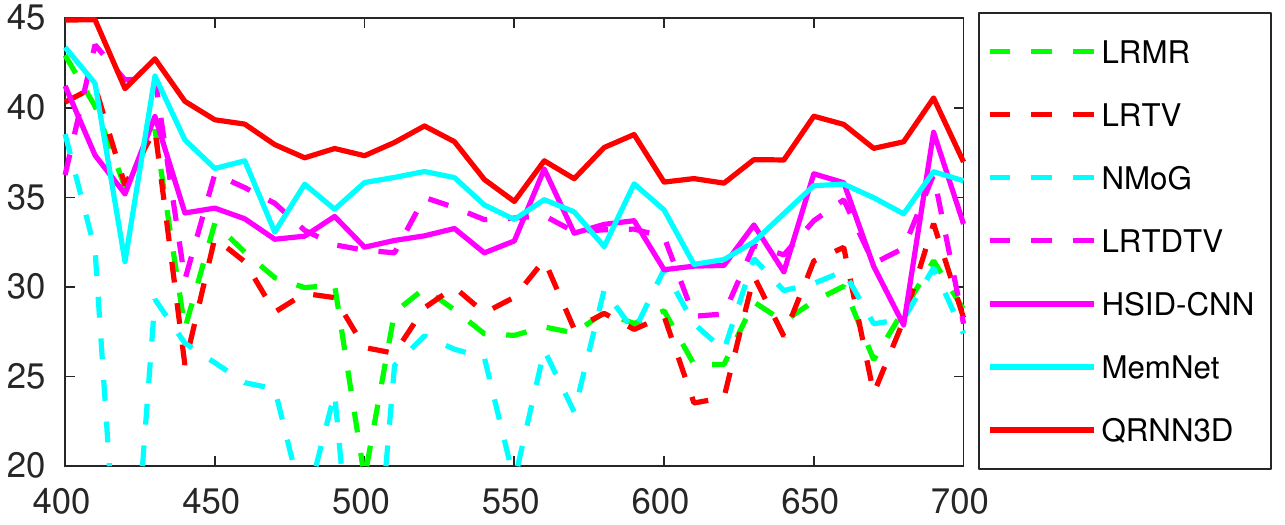}
\caption{Mixture (Case 5)}
\end{subfigure}

\caption{PSNR values across the spectrum corresponding to Gaussian and complex noise removal results in Figure \ref{fig:denoising:1} and \ref{fig:denoising:icvl_complex} respectively. } \label{fig:PSNRs}
\end{figure*}

\subsection{Experimental settings}
\subsubsection{Benchmark Datasets}
We conduct several experiments using data from ICVL hyperspectral dataset
\cite{arad_and_ben_shahar_2016_ECCV}, where 201 images were collected at
$1392\times1300$ spatial resolution over 31 spectral bands.
The simulated pseudo color image samples from this dataset are illustrated in Figure \ref{fig:icvl_rgb}.
We use 100 images for training, 5 images for validation,  while others are for testing. To enlarge the training set, 
we crop multiple overlapped volumes from training HSIs and then regard each
volume as a training sample. During cropping, 
 each volume has a spatial size of
$64 \times 64$ and a spectral size of $31$ for the purpose of preserving the
complete spectrum of an HSI.
Data augmentation schemes such as rotation and
scaling are also employed, resulting in roughly 50k training samples in total. As
for testing set, we crop the main region of each image with size of $512 \times
512 \times 31$ given the computation cost\footnote{It's unwieldy to evaluate a
  image with large size in some competing methods rather than ours, see Figure
  \ref{fig:runtime} for more detail.}.

Besides, we evaluate the robustness and flexibility of our model in remotely
sensed hyperspectral datasets including \textit{Pavia Centre}, \textit{Pavia
  University}, \textit{Indian Pines} and \textit{Urban}. 
  \textit{Pavia Centre}
and \textit{Pavia University} were acquired by the ROSIS sensor, the number of
spectral bands is 102 for \textit{Pavia Centre} and 103 for \textit{Pavia
  University}.
\textit{Indian Pines} and \textit{Urban} were gathered by 224-bands AVIRIS
sensor and 210-bands HYDICE hyperspectral system respectively. 
Both of them have been used for \textit{real} HSI denoising experiments
\cite{he2016total,wang2017hyperspectral,chang2017hyper}.

\subsubsection{Noise settings} \label{sec:noise_settings}
Real-world HSIs are usually contaminated by several different types of noise, including the most common Gaussian noise, impulse noise, dead pixels or lines, and stripes \cite{zhang2014hyperspectral, He_2015_ICCV, chen2017denoising}. 
We define five types of complex noise as follows, and the types of complex noise are referred as Case 1-5 respectively. 

\begin{enumerate}[{Case} 1:]
\item \textit{Non-i.i.d. Gaussian noise}. Entries in all bands are corrupted by
  zero-mean Gaussian noise with different intensities, randomly selected from 10
  to 70.
\item \textit{Gaussian + Stripe noise}. All bands are corrupted by non-i.i.d.
  Gaussian noise as Case 1. One third of bands (10 bands for ICVL dataset) are
  randomly chosen to add stripe noise (5\% to 15\% percentages of columns).
\item \textit{Gaussian + Deadline noise}. The noise generation process is nearly
  the same as Case 2 except the stripe noise is replaced by deadline.
\item \textit{Gaussian + Impulse noise}. Each band is contaminated by Gaussian
  noise as Case 1. One third of bands are randomly selected to add impulse
  noise with intensity ranged from 10\% to 70\%.
\item \textit{Mixture noise}. Each band is randomly corrupted by at least one
  kind of noise mentioned in Case 1-4.
\end{enumerate}

\subsubsection{Competing Methods}
We compare our method against both traditional and DL
methods in both Gaussian and complex noise cases. In general, the traditional
methods are best suited to be applied in a specific noise setting, relying on
their noise assumption. While DL methods, can be applied in various noise
setting by training multiple models to tackle miscellaneous noises. 
\textit{For the sake
of fairness, we adopt different traditional baselines in these two noise
contexts, given their noise assumptions.} 

In Gaussian noise case, we compare with several representative traditional
methods including filtering-based approaches (BM4D \cite{maggioni2013nonlocal}),
dictionary learning approach (TDL \cite{peng2014TDL}), and tensor-based
approaches (ITSReg \cite{xie2016multispectral}, LLRT \cite{chang2017hyper}). In
complex noise case, the competing traditional baselines include low-rank matrix
recovery approaches (LRMR \cite{zhang2014hyperspectral}, LRTV
\cite{he2016total}, NMoG \cite{chen2017denoising}), and low-rank tensor approach
(TDTV \cite{wang2017hyperspectral}).

For DL approaches, we compare our model with HSID-CNN
\cite{yuan2018hyperspectral}. 
  Besides, any DL method for single image denoising
can be extended to HSI denoising case (by modifying the first layer to adapt the
HSI, \ie changing $C_{in}$ from 3 to 31). For completeness, we also compare such
state-of-the-art 2D DL approach, \ie MemNet \cite{tai2017memnet} with
$C_{in}=31$ in first layer, which entails the fixed number of spectral bands. 
 Since the training setting is different between ours
and other DL approaches, we finetune/retrain their pretrained models
with our well-designed training strategy to achieve better performance in our
dataset.

\subsubsection{Network learning} 
 We develop an incremental training policy to stabilize and accelerate the training, which also avoids the network converging to a poor local minimum.  The philosophy of our training policy is simple: learning to solve tasks in an easy-to-difficult way  \cite{Ahissar2004The}. 
Networks are learned by minimizing the mean square error (MSE) between the
predicted high-quality HSI and the ground truth.  
The network parameters are initialized as in \cite{He_2015_ICCV}, and optimized using ADAM optimizer \cite{kingma2014adam} with the deep learning framework Pytorch\footnote{https://pytorch.org/} on a machine with
NVIDIA GTX 1080Ti GPU, Intel(R) Core(TM) i7-7700K CPU of 4.2GHz and 16 GB RAM.   Unlike training networks independently to tackle several different types of noise separately, we simply train two models  in both Gaussian and complex noise cases respectively. 
Our network learning goes through three stages, from the easy task of Gaussian denoising with fixed noise level, to the difficult one of complex noise removal. 
The models are incrementally trained that reuse the prior state (pretrained parameters) to maximize the training efficiency (See discussions in Section \ref{sec:efficacy-training-policy}).
We follow the previous image restoration work \cite{mao2016image} to choose hyper-parameters of learning algorithm. These values were empirically set to make network learning fast yet stable.  Specifically, the learning rate is initialized at $10^{-3}$ and decayed at epochs, where the validation performance not increases any more. Small batch size (\ie 16) is used to accelerate training at first stage, while large batch size (\ie 64) is adopted to stabilize training when tackling harder cases (\eg complex noise case).  The overview of our training procedures is shown in Table \ref{tb:training-policy}, with detailed hyper-parameter setting.

\subsubsection{Quantitative Metrics}
To give an overall evaluation, three quantitative quality indices are employed,
i.e. PSNR, SSIM \cite{wang2004image}, and SAM \cite{yuhas1993determination}.
PSNR and SSIM are two conventional spatial-based indexes, while SAM is
spectral-based. Larger values of PSNR and SSIM imply better performance, while
a smaller value of SAM suggests better performance.

\subsection{Experiments on ICVL Dataset}

\begin{figure*}[!htbp]
\centering

\begin{subfigure}[b]{.18\linewidth}
\centering
\includegraphics[height=1\linewidth,clip,keepaspectratio]{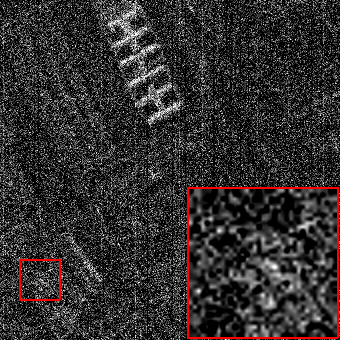}
\caption{Noisy \\\hspace{\textwidth}\centering (13.54)}
\end{subfigure}
\begin{subfigure}[b]{.18\linewidth}
\centering
\includegraphics[height=1\linewidth,clip,keepaspectratio]{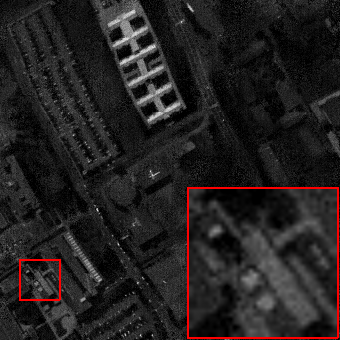}
\caption{LRMR \\\hspace{\textwidth}\centering (26.35)}
\end{subfigure}
\begin{subfigure}[b]{.18\linewidth}
\centering
\includegraphics[height=1\linewidth,clip,keepaspectratio]{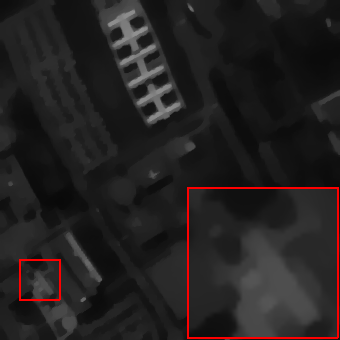}
\caption{LRTV \\\hspace{\textwidth}\centering (25.93)}
\end{subfigure}
\begin{subfigure}[b]{.18\linewidth}
\centering
\includegraphics[height=1\linewidth,clip,keepaspectratio]{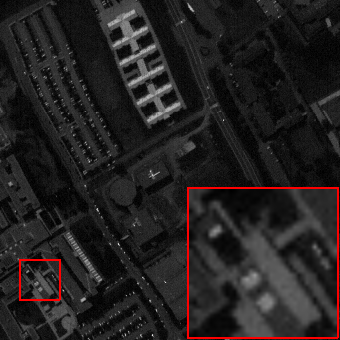}
\caption{NMoG \\\hspace{\textwidth}\centering (28.90)}
\end{subfigure}
\begin{subfigure}[b]{.18\linewidth}
\centering
\includegraphics[height=1\linewidth,clip,keepaspectratio]{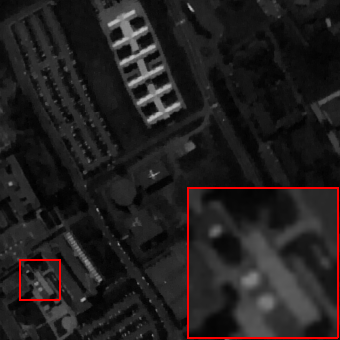}
\caption{LRTDTV \\\hspace{\textwidth}\centering (30.06)}
\end{subfigure}
\begin{subfigure}[b]{.18\linewidth}
\centering
\includegraphics[height=1\linewidth,clip,keepaspectratio]{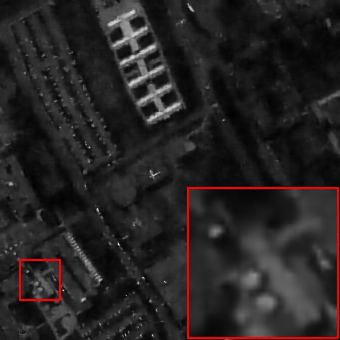}
\caption{HSID-CNN \\\hspace{\textwidth}\centering (30.14)}
\end{subfigure}
\begin{subfigure}[b]{.18\linewidth}
\centering
\includegraphics[height=1\linewidth,clip,keepaspectratio]{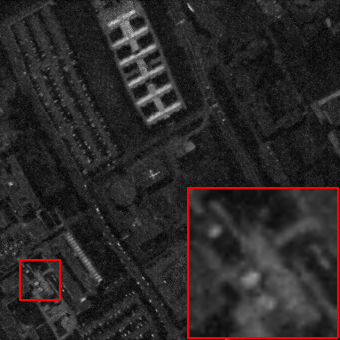}
\caption{Ours-S \\\hspace{\textwidth}\centering (29.64)}
\end{subfigure}
\begin{subfigure}[b]{.18\linewidth}
\centering
\includegraphics[height=1\linewidth,clip,keepaspectratio]{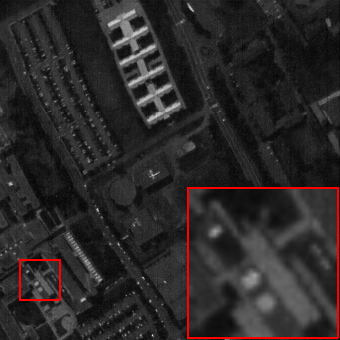}
\caption{Ours-P \\\hspace{\textwidth}\centering (31.50)}
\end{subfigure}
\begin{subfigure}[b]{.18\linewidth}
\centering
\includegraphics[height=1\linewidth,clip,keepaspectratio]{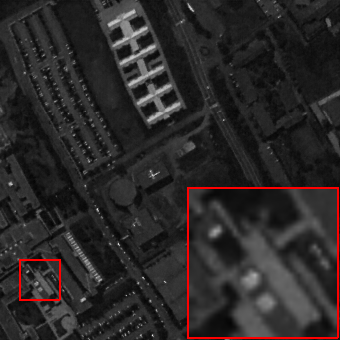}
\caption{Ours-F \\\hspace{\textwidth}\centering (34.32)}
\end{subfigure}
\begin{subfigure}[b]{.18\linewidth}
\centering
\includegraphics[height=1\linewidth,clip,keepaspectratio]{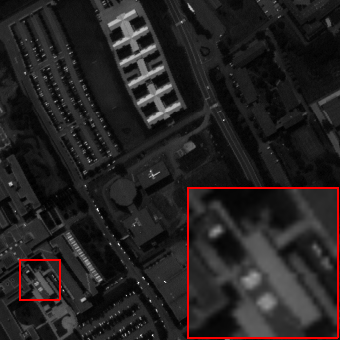}
\caption{Clean \\\hspace{\textwidth}\centering (+$\infty$)}
\end{subfigure}

\caption{Simulated complex noise removal results of PSNR (dB) at $10^{th}$ band of image in case 5 (mixture noise) on \textit{Pavia University} dataset. (\textbf{Best view on screen with zoom})}
\label{fig:denoising:pavia}

\end{figure*}

\begin{figure*}[!t]
\centering
\begin{subfigure}[b]{.16\linewidth}
\centering
\includegraphics[width=1\linewidth,clip,keepaspectratio]{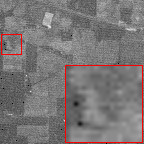}
\caption{Noisy}
\label{fig:indian:1}
\end{subfigure}
\centering
\begin{subfigure}[b]{.16\linewidth}
\centering
\includegraphics[width=1\linewidth,clip,keepaspectratio]{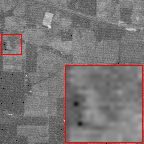}
\caption{BM4D}
\end{subfigure}
\begin{subfigure}[b]{.16\linewidth}
\centering
\includegraphics[width=1\linewidth,clip,keepaspectratio]{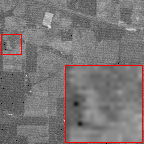}
\caption{TDL}
\end{subfigure}
\begin{subfigure}[b]{.16\linewidth}
\centering
\includegraphics[width=1\linewidth,clip,keepaspectratio]{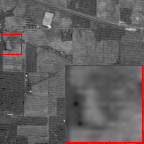}
\caption{ITSReg}
\end{subfigure}
\begin{subfigure}[b]{.16\linewidth}
\centering
\includegraphics[width=1\linewidth,clip,keepaspectratio]{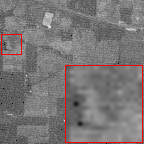}
\caption{LLRT}
\end{subfigure}
\begin{subfigure}[b]{.16\linewidth}
\centering
\includegraphics[width=1\linewidth,clip,keepaspectratio]{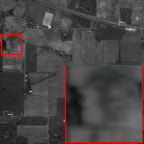}
\caption{LRMR}
\end{subfigure}
\begin{subfigure}[b]{.16\linewidth}
\centering
\includegraphics[width=1\linewidth,clip,keepaspectratio]{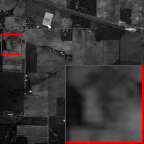}
\caption{LRTV}
\end{subfigure}
\begin{subfigure}[b]{.16\linewidth}
\centering
\includegraphics[width=1\linewidth,clip,keepaspectratio]{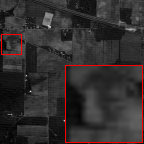}
\caption{NMoG}
\end{subfigure}
\begin{subfigure}[b]{.16\linewidth}
\centering
\includegraphics[width=1\linewidth,clip,keepaspectratio]{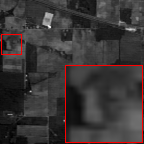}
\caption{TDTV}
\end{subfigure}
\begin{subfigure}[b]{.16\linewidth}
\centering
\includegraphics[width=1\linewidth,clip,keepaspectratio]{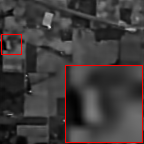}
\caption{HSID-CNN}
\end{subfigure}
\begin{subfigure}[b]{.16\linewidth}
\centering
\includegraphics[width=1\linewidth,clip,keepaspectratio]{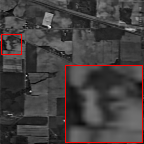}
\caption{Ours}
\end{subfigure}

\caption{Real-world unknown noise removal results at $2^{th}$ band of image on
  AVIRIS \textit{Indian Pines} dataset. (\textbf{Best view on screen with zoom})}
\label{fig:indian}
\end{figure*}

\begin{figure*}[!t]
\centering
\begin{subfigure}[b]{.16\linewidth}
\centering
\includegraphics[width=1\linewidth,clip,keepaspectratio]{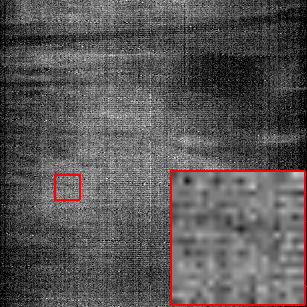}
\caption{Noisy}
\label{fig:indian:1}
\end{subfigure}
\centering
\begin{subfigure}[b]{.16\linewidth}
\centering
\includegraphics[width=1\linewidth,clip,keepaspectratio]{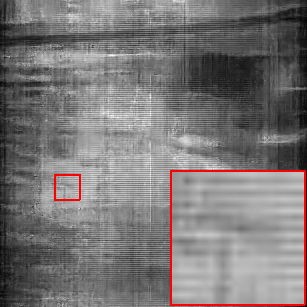}
\caption{BM4D}
\end{subfigure}
\begin{subfigure}[b]{.16\linewidth}
\centering
\includegraphics[width=1\linewidth,clip,keepaspectratio]{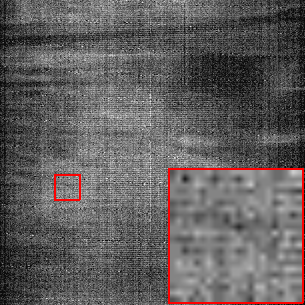}
\caption{TDL}
\end{subfigure}
\begin{subfigure}[b]{.16\linewidth}
\centering
\includegraphics[width=1\linewidth,clip,keepaspectratio]{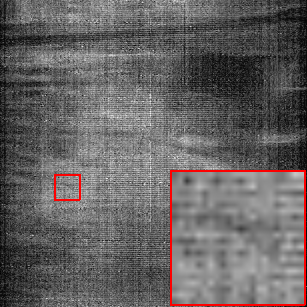}
\caption{ITSReg}
\end{subfigure}
\begin{subfigure}[b]{.16\linewidth}
\centering
\includegraphics[width=1\linewidth,clip,keepaspectratio]{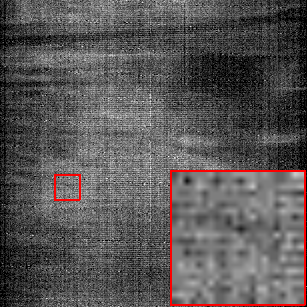}
\caption{LLRT}
\end{subfigure}
\begin{subfigure}[b]{.16\linewidth}
\centering
\includegraphics[width=1\linewidth,clip,keepaspectratio]{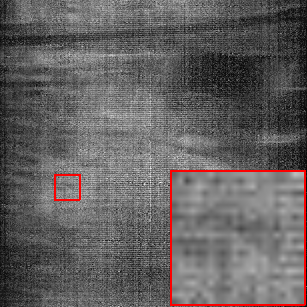}
\caption{LRMR}
\end{subfigure}
\begin{subfigure}[b]{.16\linewidth}
\centering
\includegraphics[width=1\linewidth,clip,keepaspectratio]{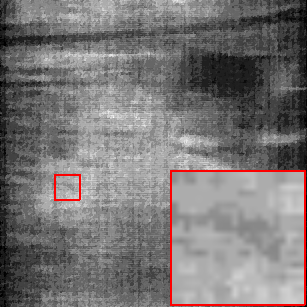}
\caption{LRTV}
\end{subfigure}
\begin{subfigure}[b]{.16\linewidth}
\centering
\includegraphics[width=1\linewidth,clip,keepaspectratio]{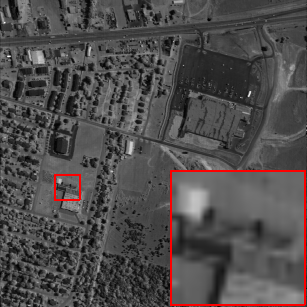}
\caption{NMoG}
\end{subfigure}
\begin{subfigure}[b]{.16\linewidth}
\centering
\includegraphics[width=1\linewidth,clip,keepaspectratio]{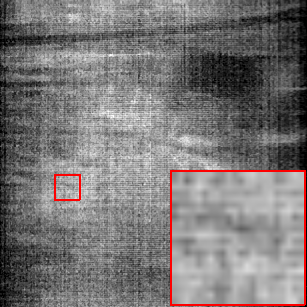}
\caption{TDTV}
\end{subfigure}
\begin{subfigure}[b]{.16\linewidth}
\centering
\includegraphics[width=1\linewidth,clip,keepaspectratio]{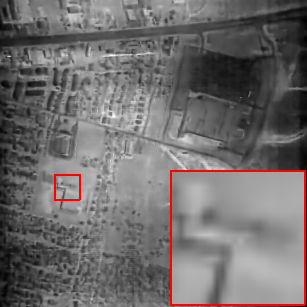}
\caption{HSID-CNN}
\end{subfigure}
\begin{subfigure}[b]{.16\linewidth}
\centering
\includegraphics[width=1\linewidth,clip,keepaspectratio]{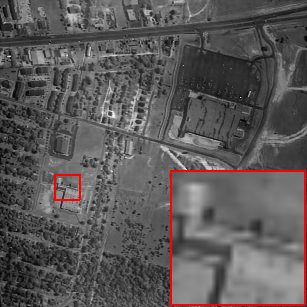}
\caption{Ours}
\end{subfigure}

\caption{Real-world unknown noise removal results at $107^{th}$ band of image on
  HYDICE \textit{Urban} dataset. (\textbf{Best view on screen with zoom})}
\label{fig:urban}
\end{figure*}

\begin{table*}[!htbp]
	\centering
	\caption{Quantitative results of different methods under several noise levels on ICVL dataset. "Blind" suggests each sample is corrupted by Gaussian noise with unknown $\sigma$ (ranged from 30 to 70). }
	\setlength{\tabcolsep}{1mm}{
	\begin{tabular}{|C{.085\linewidth}|C{.085\linewidth}|C{.085\linewidth}|C{.085\linewidth}|C{.085\linewidth}|C{.085\linewidth}|C{.085\linewidth}|C{.085\linewidth}|C{.085\linewidth}|C{.085\linewidth}|}
		\hline
		\multirow{3}{*}{Sigma} & \multirow{3}{*}{Index} & \multicolumn{8}{c|}{Methods} \\ \cline{3-10}
		& & Noisy & BM4D & TDL & ITSReg & LLRT & HSID-CNN & MemNet & Ours \\
		& & & \cite{maggioni2013nonlocal} & \cite{peng2014TDL} & \cite{xie2016multispectral} & \cite{chang2017hyper} & \cite{yuan2018hyperspectral}& \cite{tai2017memnet} &  \\ \hline
\multirow{3}{*}{30} &PSNR&$18.59$&$38.45$&$40.58$&$41.48$&$41.99$&$38.70$&$41.45$&$\bm{42.28}$\\\cline{2-10}
&SSIM&$0.110$&$0.934$&$0.957$&$0.961$&$0.967$&$0.949$&$0.972$&$\bm{0.973}$\\\cline{2-10}
&SAM&$0.807$&$0.126$&$0.062$&$0.088$&$\bm{0.056}$&$0.103$&$0.065$&$0.061$\\\hline
\multirow{3}{*}{50} &PSNR&$14.15$&$35.60$&$38.01$&$38.88$&$38.99$&$36.17$&$39.76$&$\bm{40.23}$\\\cline{2-10}
&SSIM&$0.046$&$0.889$&$0.932$&$0.941$&$0.945$&$0.919$&$0.960$&$\bm{0.961}$\\\cline{2-10}
&SAM&$0.991$&$0.169$&$0.085$&$0.098$&$0.075$&$0.134$&$0.076$&$\bm{0.072}$\\\hline
\multirow{3}{*}{70} &PSNR&$11.23$&$33.70$&$36.36$&$36.71$&$37.36$&$34.31$&$38.37$&$\bm{38.57}$\\\cline{2-10}
&SSIM&$0.025$&$0.845$&$0.909$&$0.923$&$0.930$&$0.886$&$\bm{0.946}$&$0.945$\\\cline{2-10}
&SAM&$1.105$&$0.207$&$0.105$&$0.112$&$0.087$&$0.161$&$0.088$&$\bm{0.087}$\\\hline
\multirow{3}{*}{Blind} &PSNR&$17.34$&$37.66$&$39.91$&$40.62$&$40.97$&$37.80$&$40.70$&$\bm{41.50}$\\\cline{2-10}
&SSIM&$0.114$&$0.914$&$0.946$&$0.953$&$0.956$&$0.935$&$0.966$&$\bm{0.967}$\\\cline{2-10}
&SAM&$0.859$&$0.143$&$0.072$&$0.087$&$\bm{0.064}$&$0.116$&$0.070$&$0.066$\\\hline
	\end{tabular}}
	\label{tb:denosing:1}
\end{table*}

\begin{table*}[!htbp]
	\centering

	\caption{Quantitative results of different methods in five complex noise cases on ICVL dataset.}
	\setlength{\tabcolsep}{1mm}{
	\begin{tabular}{|C{.085\linewidth}|C{.085\linewidth}|C{.085\linewidth}|C{.085\linewidth}|C{.085\linewidth}|C{.085\linewidth}|C{.085\linewidth}|C{.085\linewidth}|C{.085\linewidth}|C{.085\linewidth}|}
		\hline
		\multirow{3}{*}{Case} & \multirow{3}{*}{Index} & \multicolumn{8}{c|}{Methods} \\ \cline{3-10}
		& & Noisy & LRMR & LRTV & NMoG & TDTV & HSID-CNN & MemNet & Ours \\
		& & & \cite{zhang2014hyperspectral} & \cite{he2016total} & \cite{chen2017denoising} & \cite{wang2017hyperspectral} & \cite{yuan2018hyperspectral}& \cite{tai2017memnet} & \\ \hline
\multirow{3}{*}{1} &PSNR&$18.25$&$32.80$&$33.62$&$34.51$&$38.14$&$38.40$&$38.94$&$\bm{42.79}$\\\cline{2-10}
&SSIM&$0.168$&$0.719$&$0.905$&$0.812$&$0.944$&$0.947$&$0.949$&$\bm{0.978}$\\\cline{2-10}
&SAM&$0.898$&$0.185$&$0.077$&$0.187$&$0.075$&$0.095$&$0.091$&$\bm{0.052}$\\\hline
\multirow{3}{*}{2} &PSNR&$17.80$&$32.62$&$33.49$&$33.87$&$37.67$&$37.77$&$38.57$&$\bm{42.35}$\\\cline{2-10}
&SSIM&$0.159$&$0.717$&$0.905$&$0.799$&$0.940$&$0.942$&$0.945$&$\bm{0.976}$\\\cline{2-10}
&SAM&$0.910$&$0.187$&$0.078$&$0.265$&$0.081$&$0.104$&$0.095$&$\bm{0.055}$\\\hline
\multirow{3}{*}{3} &PSNR&$17.61$&$31.83$&$32.37$&$32.87$&$36.15$&$37.65$&$38.15$&$\bm{42.23}$\\\cline{2-10}
&SSIM&$0.155$&$0.709$&$0.895$&$0.797$&$0.930$&$0.940$&$0.945$&$\bm{0.976}$\\\cline{2-10}
&SAM&$0.917$&$0.227$&$0.115$&$0.276$&$0.099$&$0.102$&$0.096$&$\bm{0.056}$\\\hline
\multirow{3}{*}{4} &PSNR&$14.80$&$29.70$&$31.56$&$28.60$&$36.67$&$35.00$&$35.93$&$\bm{39.23}$\\\cline{2-10}
&SSIM&$0.114$&$0.623$&$0.871$&$0.652$&$0.935$&$0.899$&$0.907$&$\bm{0.945}$\\\cline{2-10}
&SAM&$0.926$&$0.311$&$0.242$&$0.486$&$\bm{0.094}$&$0.174$&$0.126$&$0.109$\\\hline
\multirow{3}{*}{5} &PSNR&$14.08$&$28.68$&$30.47$&$27.31$&$34.77$&$34.05$&$35.16$&$\bm{38.25}$\\\cline{2-10}
&SSIM&$0.099$&$0.608$&$0.858$&$0.632$&$0.919$&$0.888$&$0.903$&$\bm{0.938}$\\\cline{2-10}
&SAM&$0.944$&$0.353$&$0.287$&$0.513$&$0.113$&$0.181$&$0.130$&$\bm{0.107}$\\\hline
	\end{tabular}}
	\label{tb:denoising:2}
\end{table*}

\begin{table*}[!htbp]
	\centering
	\caption{ Quantitative results of different methods in mixture noise case on
      \textit{Pavia University} dataset. 
	"Ours-S" is our trained-from-scratch model which is only trained on \textit{Pavia Centre} dataset; "Ours-P" denotes our pretrained model which is only trained on ICVL dataset;
	"Ours-F" indicates our fine-tuned model which is pretrained on ICVL dataset, and then is fine-tuned on \textit{Pavia Centre} dataset. }
	\setlength{\tabcolsep}{1mm}{
	\begin{tabular}{|C{.085\linewidth}|C{.085\linewidth}|C{.085\linewidth}|C{.085\linewidth}|C{.085\linewidth}|C{.085\linewidth}|C{.085\linewidth}|C{.085\linewidth}|C{.085\linewidth}|C{.085\linewidth}|C{.085\linewidth}|}
		\hline
		\multirow{3}{*}{Index} & \multicolumn{9}{c|}{Methods} \\ \cline{2-10}
	    &  Noisy & LRMR & LRTV & NMoG & TDTV &  HSID-CNN &Ours & Ours & Ours \\
	  & & \cite{zhang2014hyperspectral} & \cite{he2016total} & \cite{chen2017denoising} & \cite{wang2017hyperspectral} & \cite{yuan2018hyperspectral} & S & P & F \\ \hline
PSNR&$13.54$&$26.35$&$25.93$&$28.90$&$30.06$&$30.14$&$29.64$&$31.50$&$\bm{34.32}$\\\hline
SSIM&$0.161$&$0.660$&$0.676$&$0.781$&$0.819$&$0.805$&$0.892$&$0.866$&$\bm{0.925}$\\\hline
SAM&$0.896$&$0.406$&$0.359$&$0.388$&$0.239$&$0.142$&$0.166$&$0.127$&$\bm{0.093}$\\\hline 
	\end{tabular}}
	\label{tb:pavia}
\end{table*}

\subsubsection{Denoising in Gaussian Noise Case}
Zero mean additive white Gaussian noises with different variance are added to
generate the noisy observations. 
The model trained at the end of stage 2 (epoch 50) is used to tackle all different
levels of corruption\footnote{We do not train multiple networks to tackle
  different noise intensities respectively. Instead, only one single network is
  trained using training sample with various noise intensities. }. 
  Figure
\ref{fig:denoising:1} shows the denoising results under noise level $\sigma =
50$. It can be easily observed that the image restored by our method is capable
of properly removing the Gaussian noise while finely preserving the structure
underlying the HSI. Traditional methods like BM4D and TDL
introduce evident artifacts to some areas. Other methods suppress
the noise better, but still lose some fine-grained details and produce
relatively low-quality results compared with ours. The qualitative assessment
results are listed in Table \ref{tb:denosing:1}. Compared with all competing
methods, the QRNN3D achieves better performance in most 
qualitative/quantitative assessments, further confirming the high fidelity of
our method.

\subsubsection{Denoising in Complex Noise Case} \label{sec:complex} Five types
of the complex noise are added to generate noisy samples. In brief, cases 1-5
represent \textit{non-i.i.d Gaussian noise, Gaussian + stripes, Gaussian +
  deadline, Gaussian + impulse, and mixture of them} respectively (see Section
\ref{sec:noise_settings} for more details). 
Like Gaussian noise case, a single model trained at the end of stage 3 (epoch 100) is utilized to dealing with case 1-5 simultaneously. 
It's worth noting
that each sample in our training set is corrupted by one of noise types (\ie
cases 1-4), while in case 5, each testing sample suffers from multiple types of
noise, not contained in the training set. We show the qualitative and
quantitative results in Figure \ref{fig:denoising:icvl_complex} and Table
\ref{tb:denoising:2} respectively, which show our QRNN3D significantly
outperforms the other methods. Furthermore, the results in mixture noise case
exhibit the strong generalization of our model since the mixture noise is not
seen by our model in the training stage.

In Figure \ref{fig:denoising:icvl_complex}, the observation images are
corrupted by miscellaneous complex noises. Low-rank matrix recovery methods, i.e. LRMR
and LRTV, holding the assumption that the clean HSI lies in low-rank subspace
from the spectral perspective, successfully remove great mass of noise, but at a
cost of losing fine details. Our QRNN3D eliminates miscellaneous noises to a
great extent, while more faithfully preserving the fine-grained structure of
original image (\eg the texture of road in the second photo of Figure
\ref{fig:denoising:icvl_complex}) than top-performing traditional low-rank
tensor approach TDTV and other DL methods. Figure \ref{fig:PSNRs} shows the PSNR
value of each bands in these HSIs. It can be seen that the PSNR values of all
bands obtained by Our QRNN3D are obviously higher than those compared methods.

\subsection{Experiments on Remotely Sensed Images}

\subsubsection{Synthetic Data}
Here, we conduct experiments on \textit{Pavia University} in mixture noise case.
Given the similarity between \textit{Pavia Centre} and \textit{Pavia
  University}, the model is first trained from scratch only on \textit{Pavia
  Centre}. It can be seen our train-from-scratch model (Ours-S in Table
\ref{tb:pavia}) performs undesirable, even compared with traditional method TDTV
(29.64 v.s. 30.06).

Nevertheless, our method utilizes QRU3D, which makes it can
be naturally used for input data with various number of bands. On the basis of
this flexibility, we directly apply our model pretrained on ICVL dataset (in
complex noise case) to \textit{Pavia University}. 
Although the \textit{Pavia University} is recorded with a spectral curve totally
distinct from ICVL dataset, our model called Ours-P performs much better than
all compared methods\footnote{The result of HSID-CNN is also obtained by its
  pretrained model on ICVL dataset under complex noise case. The learned
  MemNet cannot be useful for the data with different bands and its results are
  not provided in Table \ref{tb:pavia}.}, which strongly
verifies the robustness of our method. 

Furthermore, we employ small pieces of samples from \textit{Pavia Center} to
fine-tune the model only learned from ICVL dataset. This learned model (Ours-F
in Table \ref{tb:pavia}) significantly boosts the performance. The visual
comparison is provided in Figure \ref{fig:denoising:pavia}. Interestingly, the
Gaussian-like residuals are still visible in Ours-S model, while Ours-P model
suffers from stripes. Ours-F model combines the strengths of the two models,
yielding clear and clean result. This seems to indicate the knowledge from ICVL
dataset is complementary to one from \textit{Pavia Centre} dataset, so that the
transfer learning enabled by flexibility will bring great benefits in
performance.

\subsubsection{Real-world Noisy Data}
We also verify our model in real-world noisy HSI \textit{Indian Pines} and
\textit{Urban} without corresponding ground truth. It can be observed in Figure
\ref{fig:indian} and Figure \ref{fig:urban} that terrible atmosphere and water
absorption obstruct the view to the real scenario, severely degrading the
quality of images. The Gaussian denoising methods, \eg BM4D, TDL, cannot
accurately estimate the underlying clean image due to the non-Gaussian noise
structure. Our QRNN3D successfully tackles this unknown noise, and
produces sharper and clearer result than others, consistently demonstrating
the robustness and flexibility of our model.

\begin{table}[!t] 
\centering
\caption{Ablations on ICVL HSI Gaussian denoising (under noise level $\sigma=50$). We evaluate the results by PSNR (dB), running Time (sec) and the number of parameters (Params) of these networks. All running times are measured on a Nvidia GTX 1080Ti by processing an HSI with size of 512 $\times$ 512 $\times$ 31. 
  Direction of network is denoted by initials, \ie U: unidirectional; B:
  bidirectional;  A: alternating directional, Our benchmark network is indicated by \textbf{boldface}. The results of MemNet are also provided as an additional reference.} 
\begin{tabular}{lccc} 
		\toprule
		Model & PSNR (dB) & Time (s) & Params (\#)  \\ 
		\midrule
		MemNet & 39.76 & 0.88 & 2.94M \\ \hline
		\midrule
		QRU2D & 38.63 & 0.60  & 0.29M \\ \hline
		WQRU2D & 39.82 & 1.16 & 0.88M \\ \hline
		C3D & 36.83 &  0.56 & 0.43M \\ \hline
		WC3D & 40.00 &  0.93 & 1.72M \\ \hline		
		\textbf{QRU3D} & 40.23 & 0.74 & 0.86M \\
		\midrule	
		\midrule		
		U & 40.07 & 0.75 &  0.86M \\ \hline		
		B & 40.26 &  1.26 & 1.72M \\ \hline
		\textbf{A} & 40.23 & 0.74 & 0.86M \\
		\bottomrule
\end{tabular} 
\label{tb:ablation}
\end{table}

\begin{figure}[!t]
\centering
\begin{subfigure}[b]{.45\linewidth}
\centering
\includegraphics[width=1\linewidth,clip,keepaspectratio]{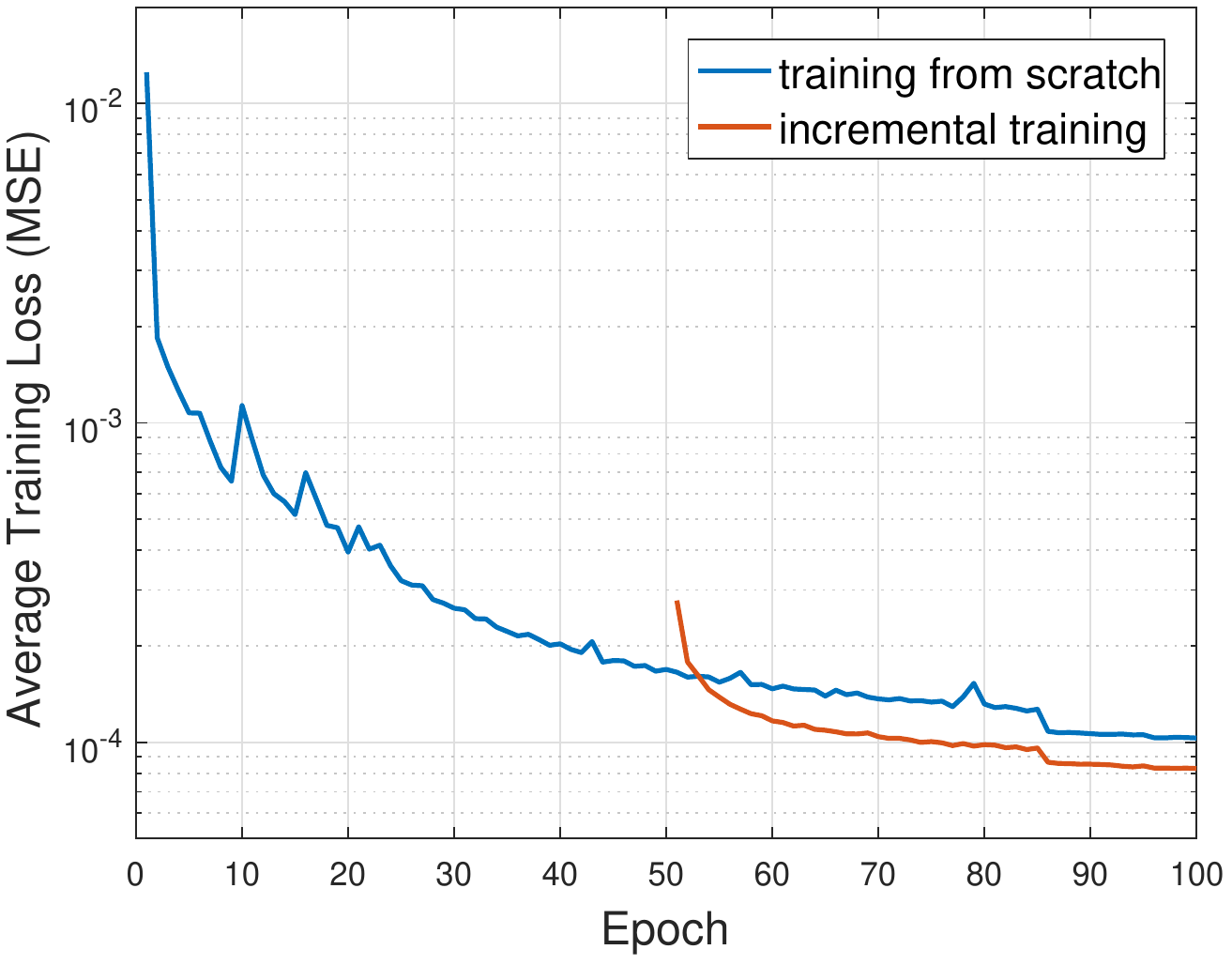}
\end{subfigure}
\begin{subfigure}[b]{.45\linewidth}
  \centering
\includegraphics[width=1\linewidth,clip,keepaspectratio]{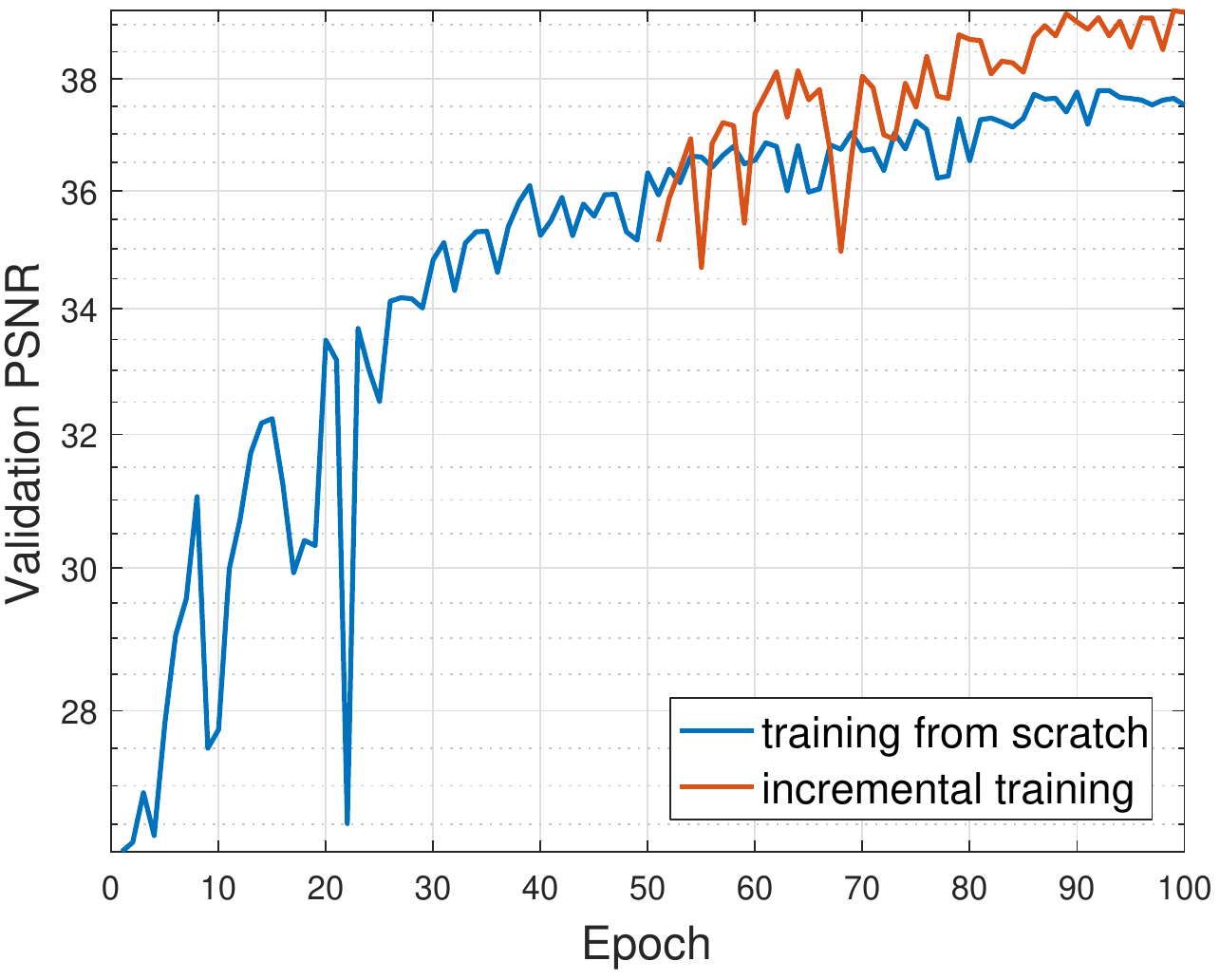}
\end{subfigure}
\caption{Average training loss (Left) and Validation PSNR (Right) of QRNN3D for complex noise removal. We show the results of the model trained from scratch, and the one that reuses the pretrained parameters in Gaussian denoising (incremental training).}
\label{fig:incremental-training}
\end{figure}

\section{Discussion and Analysis} 
\label{sec:understand}

\begin{figure*}[!htbp]
\centering
\begin{subfigure}[b]{.33\linewidth}
\centering
\includegraphics[width=1\linewidth]{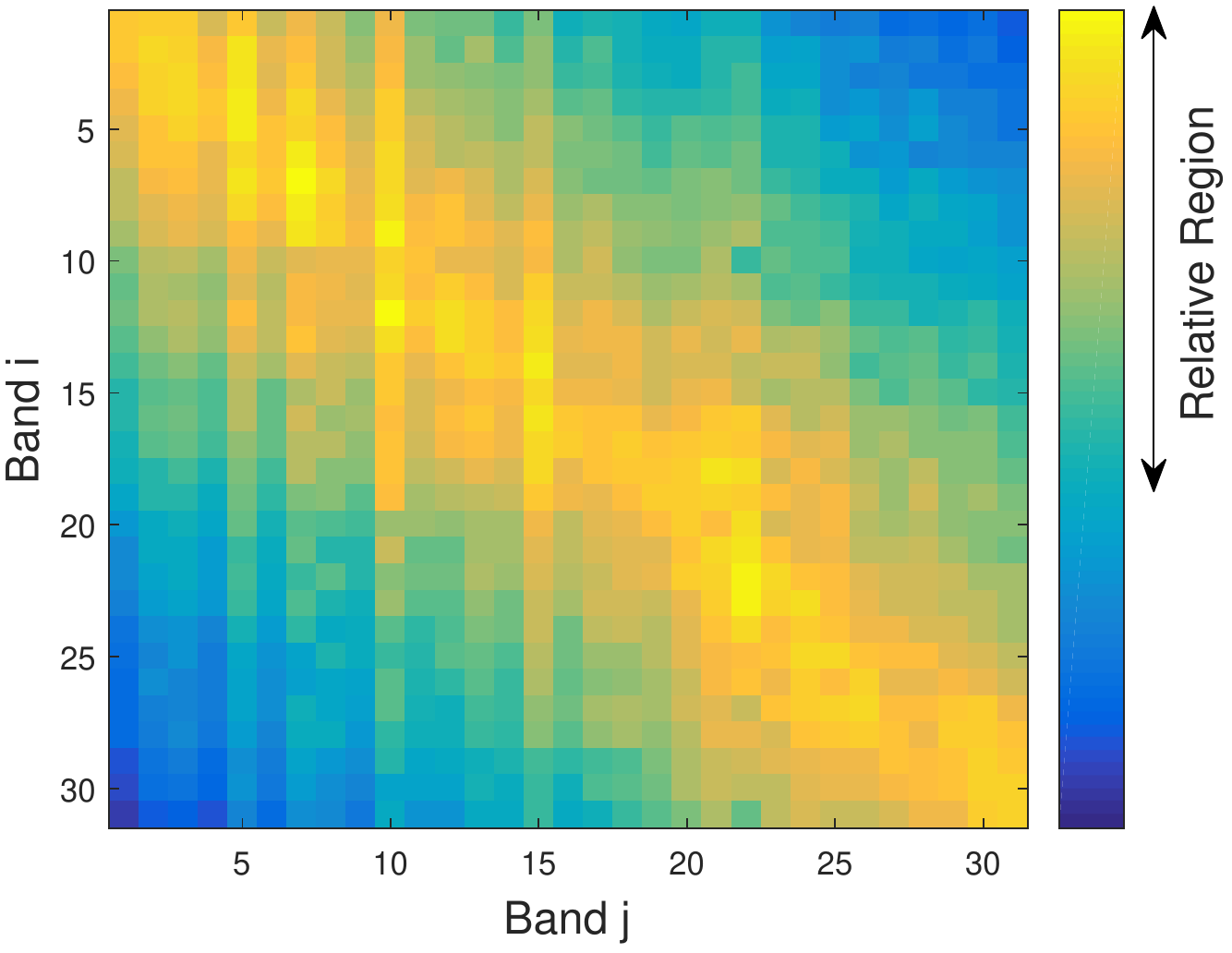}

\caption{}
\label{fig:gcs}
\end{subfigure}
\begin{subfigure}[b]{.32\linewidth}
\centering
\includegraphics[width=1\linewidth]{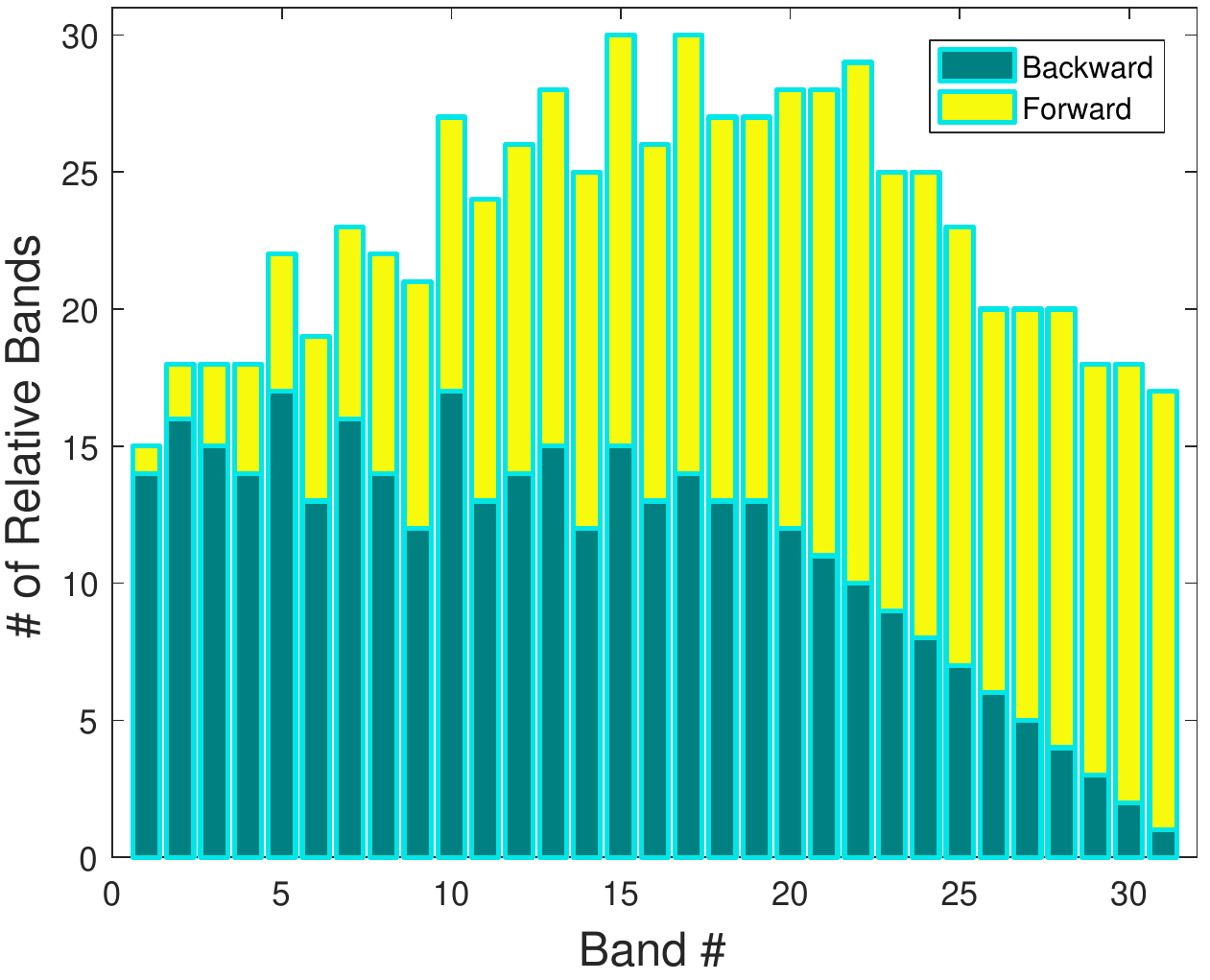}

\caption{}
\label{fig:gcs_bar}
\end{subfigure}
\begin{subfigure}[b]{.33\linewidth}
\centering
\includegraphics[width=1\linewidth]{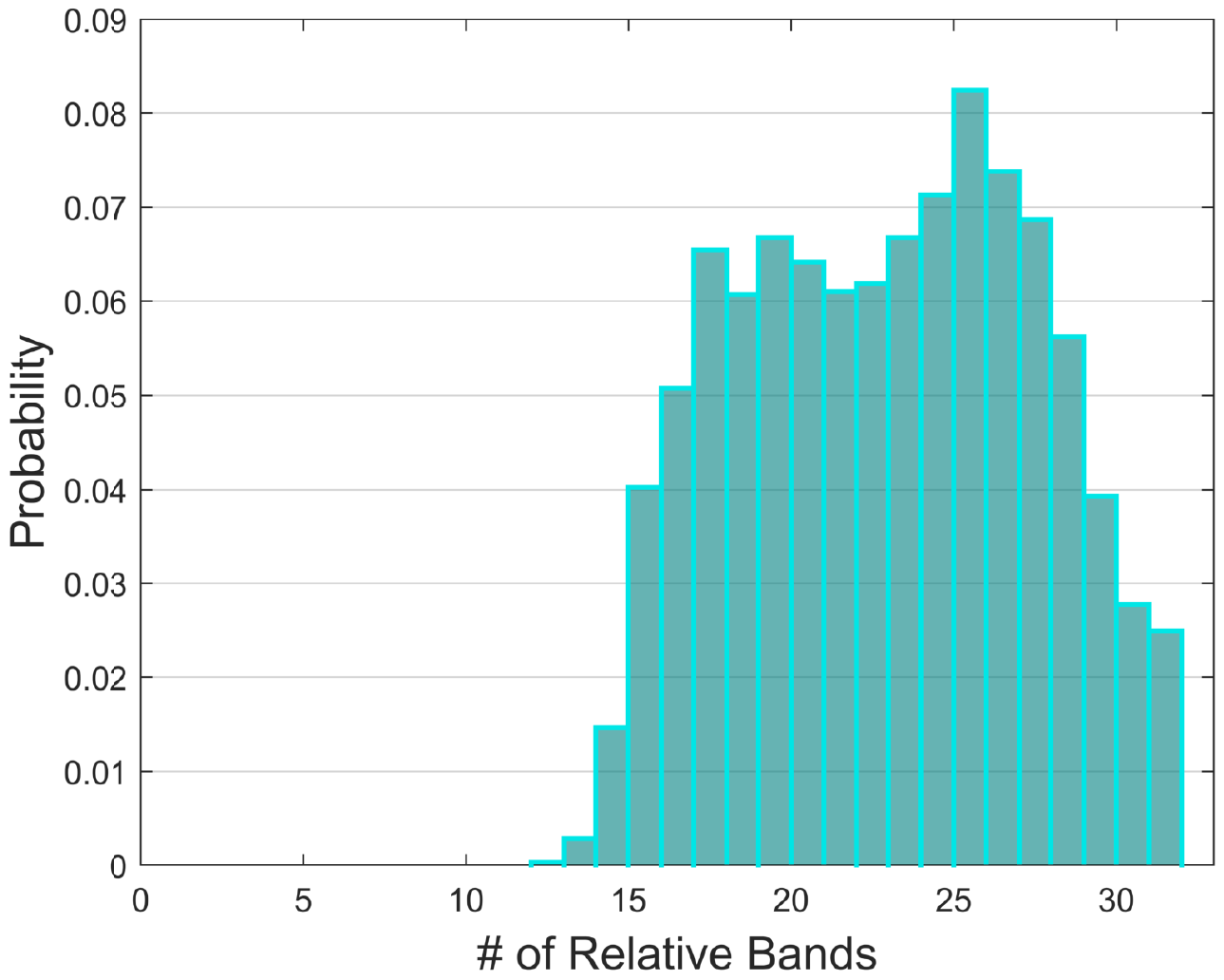}
\caption{}
\label{fig:gcs_dist}
\end{subfigure}
\caption{(a) The captured GCS in a bidirectional QRU3D layer. (b) The number of
  relative bands for output of each band. Band $i$ defined as an "relative band" for band
  $j$ means it will produce at least 10\% perturbation (\ie $GCS_{ij} \ge 0.1
  \|\bm{1}\|_F$, where $\bm{1}$ has same size as $\bm{\mathrm{h}}_{j}$ with all
  entries equal to 1) to the output if discarded. Forward/Backward denotes the
  direction of dependency. (\ie $i < j$ for forward direction). (c) The
  empirical distribution of the number of relative bands.}
\end{figure*}

In this section, we provide a broad discussion and analysis of QRNN3D to facilitate understanding of where its great performance comes from. We first demonstrate the efficacy of our incremental training policy, then analyze the functionality of each network component in QRNN3D (i.e. 3D convolution, quasi-recurrent pooling, alternating-directional structure). The selection of network hyper-parameters is followed. 
The visualization method (and results) of GCS knowledge in QRU3D are presented in final.

\subsection{Efficacy of Incremental Training Policy} \label{sec:efficacy-training-policy}
The key idea of our training policy lies at the fact that knowledge can be efficiently learned in an easy-to-difficult way \cite{Ahissar2004The}. 
Our training policy enables reusing prior learned knowledge (pretrained parameters), which significantly stabilizes and accelerates the whole training process.  As an example, we show the optimization curves with and without reusing the pretrained parameters when training the model in complex noise case. As shown in Figure \ref{fig:incremental-training}, training from scratch renders the optimization slow, instable and converge to a poor local minimum, in contrast to training with a good initialization in our incremental learning policy. 

\subsection{Component Analysis in QRNN3D}

To thoroughly verify the functionality of each component in our QRNN3D,
comprehensive ablation experiments are conducted on HSI Gaussian denoising task
on ICVL dataset. We focus on the components associated with HSI modeling and
domain knowledge embedding, and study the best trade-off between performance and
computational burden. The evaluation measures include PSNR, running
time and total number of parameters of network.  

We choose our encoder-decoder QRNN3D 
as the benchmark. For fair comparison, same network architecture is used except
the modification in the investigated component. Ablation results are exhibited
in Table \ref{tb:ablation} and analyzed in the following.

\subsubsection{Subcomponents Investigation}
Table \ref{tb:ablation} investigates the effect of subcomponents (\ie 3D
convolution and quasi-recurrent pooling function) in QRU3D. QRU3D is the basic
building block of our QRNN3D. 
 In the experiments, four
variants of this basic block are tested, \ie QRU2D,WQRU2D, C3D and WC3D. 

QRU2D is
instantiated by replacing the 3D convolution by 2D convolution (implemented by
simply setting the kernel size to $3\times 3 \times 1$). Drastic performance
losing (\ie -1.6 dB) can be observed in Table \ref{tb:ablation}, meaning
ignoring the structural spectral correlation would severely impact the model
capacity. 

WQRU2D is formed by a wider QRU2D model whose number of parameters is comparable to  QRU3D. Nevertheless, It can be observed that the QRU3D still outperforms the WQRU2D, even with less computation cost,  which suggests the higher efficiency of 3D convolution against the 2D approach towards HSI modeling.   

C3D is constructed by removing the quasi-recurrent pooling (and the
associated neural gates), definitely a residual encoder-decoder 3D convolutional neural network.
We find lack of mechanism to model the GCS, would degrade the performance by a
large margin (-3.4 dB). 

WC3D is built by a wider C3D model with more parameters (four times as much as the C3D model). It can be seen the PSNR
of QRU3D is 40.23 dB, higher than the WC3D's 40.00 dB. This suggests that the
improvement of quasi-recurrent pooling is \textit{not} just because it adds width
to the C3D model. Besides, 
 the QRU3D has only $\sim 50\%$ parameters and $\sim 80\%$ running time
of the WC3D model and is also narrower. This comparison shows that the
improvement from quasi-recurrent pooling is complementary to going wider in
standard ways.

\subsubsection{Direction of Network}
Table \ref{tb:ablation} also shows the results of
different directional structures denoted by initials (\eg U for unidirectional,
e.t.c.). Without considering backward spectral dependency, the unidirectional
architecture performs worst. After eliminating the causal dependency, both
alternating directional and bidirectional architectures significantly exceed the
unidirectional one, and achieve similar performance (40.26 v.s. 40.23).
Nevertheless, the bidirectional version requires much larger memory footprint
than ours alternating directional structure, indicating the alternating
directional structure can be used as a lightweight alternative to the typical
bidirectional one.

\begin{table}[!t] 
\centering
\caption{Network hyper-parameter selection on ICVL HSI Gaussian denoising (under noise level $\sigma=50$) through a small grid search. We evaluate the results by PSNR (dB), running Time (sec) and the number of parameters
(Params) of these networks. The selected parameters are indicated by \textbf{boldface}.} 
\begin{tabular}{|C{.15\linewidth}|C{.15\linewidth}|C{.15\linewidth}|C{.15\linewidth}|C{.15\linewidth}|} 
		\hline
		Depth & Width & PSNR (dB) & Time (s) & Params (\#)  \\ 
		\hline
		10 & \multirow{3}{*}{\textbf{16}} & 39.85 & 0.68 & 0.42M \\ \cline{3-5} \cline{1-1}
		\textbf{12} &  & 40.23 & 0.74 & 0.86M \\ \cline{3-5} \cline{1-1}
		14 &  &  39.52 & 0.80 & 1.30M \\ \hline 
		
		 \multirow{3}{*}{\textbf{12}} & 12 & 39.82  &  0.62 & 0.48M \\ \cline{3-5} \cline{2-2}
		 & \textbf{16} & 40.23 & 0.74 & 0.86M\\ \cline{3-5} \cline{2-2}
		 & 20 & 40.01 & 1.18 & 1.34M \\
		\hline
\end{tabular} 
\label{tb:network-hyperparameter}
\end{table}

\subsection{Network Hyperparameter Selection}
Our principle of network hyper-parameter selection is to make it compact yet work. Table \ref{tb:network-hyperparameter} shows the results of hyper-parameter selection on Gaussian denoising task through a small grid search, where we select the depth and width of our QRNN3D considering the best tradeoff between performance and computation overload.  

Nonetheless, we note the major goal of this work is to introduce a novel building block, specially tailored to model HSI. Such building block can be naturally inserted into any network topology, not restricted to the encoder-decoder network used in this paper.  We mainly show the effectiveness of our proposed building block and don't pursue higher performance via exhaustive search of other configurations. We have demonstrated state-of-the-art performance of our QRNN3D without heavy engineering effort on network hyper-parameter selection. Our current hyper-parameter setting might not be perfect, and the performance could be boosted potentially by parameter tuning, though this is not a major focus of this paper. 

\subsection{Visualizing GCS Knowledge}

To visualize the captured GCS knowledge in
QRNN3D, we first unfold 
the Equation \eqref{eq:f} and obtain
\begin{equation}
\bm{\mathrm{h}}_{j} = \sum_{i=1}^{j} \Phi_j (  \bm{\mathrm{z}}_{i}) \quad \forall i, j \in [1, B], \, i \le j,
\end{equation}
where 
$\Phi_j (  \bm{\mathrm{z}}_{i}) = \bm{\mathrm{f}}_j \odot \bm{\mathrm{f}}_{j-1}  \cdots \odot \bm{\mathrm{f}}_{i+1} \odot (1-\bm{\mathrm{f}}_{i}) \odot \bm{\mathrm{z}}_{i}$.

We define the $GCS_{ij}$ by the degree of $\bm{\mathrm{z}}_{i}$'s contribution
to $\bm{\mathrm{h}}_{j}$ under Frobenius norm measure, \ie 
\begin{equation}
  GCS_{ij} =  \|  \Phi_j (  \bm{\mathrm{z}}_{i}) / \bm{\mathrm{h}}_{j} \|_F  \label{eq:GCS},
\end{equation} 
where $/$ denotes element-wise division. It also implies the band $i$'s effect on
band $j$. The captured GCS in each QRU3D layer can be calculated through a single
inference pass by using Equation \eqref{eq:GCS}. 
To completely visualize GCS\footnote{in a forward (backward) QRU3D, the
  captured GCS is an upper (lower) triangular matrix}, we choose the first
bidirectional QRU3D for such analysis\footnote{The body of QRNN3D is equipped with
  the alternating directional structure, while in head and tail, the
  bidirectional directional structure is employed to avoid directional bias.}. 
Figure \ref{fig:gcs} exhibits the captured GCS of a random selected HSI, showing
the output of each band would be highly affected by the whole spectrum. Figure
\ref{fig:gcs_bar} illustrates the number of relative bands for output of each
band. It can be seen that 15th to 17th bands ($\bm{\mathrm{h}}_{15},
\bm{\mathrm{h}}_{17}$) are deeply correlated to almost all bands
($\bm{\mathrm{Z}}$). Figure \ref{fig:gcs_dist} summarizes this statistics of all
testing images on ICVL. It shows that a randomly selected band would be
typically related to at least 15 bands (31 in total), meaning the GCS is
effectively utilized by our model and our method can also automatically determine the
most relative bands across global spectra. 


\section{Conclusions} \label{sec:conclusion}

In this paper, we have proposed an alternating directional 3D quasi-recurrent
neural network for hyperspectral image denoising. Our main contribution is the
novel use of 3D convolution subcomponent, quasi-recurrent pooling function, and
alternating directional scheme for efficient spatio-spectral dependency
modeling. We have applied our model to resolve HSI denoising beyond the
Gaussian, especially in the very challenging real-world complex noise case, and
achieve better performance and faster speed. We also show our model pretrained
on ICVL dataset can be directly utilized to tackle remotely sensed images which
is infeasible in most of existing DL approaches for the HSI modeling.

In addition, the visualized results for global correlation along spectrum (GCS)
in our 3D quasi-recurrent unit (QRU3D) further experimentally convinces the GCS
is effectively exploited by our model. It's also worth investigating the proposed
QRU3D in other image sequence modeling tasks in future.

{\small
\bibliographystyle{ieee}

}

\end{document}